\documentclass[journal]{IEEEtran}
\usepackage{amsmath,amsfonts}
\usepackage{algorithmic}
\usepackage{multirow}
\usepackage{color} 
\usepackage{array}
\usepackage[caption=false,font=normalsize,labelfont=sf,textfont=sf]{subfig}
\usepackage{textcomp}
\usepackage{stfloats}
\usepackage{url}
\usepackage{verbatim}
\usepackage{graphicx}
\usepackage{cite}
\begin{document}
\title{TMFNet: Two-Stream Multi-Channels Fusion Networks for Color Image Operation \\Chain Detection }

\author{Yakun Niu, Lei Tan, Lei Zhang and Xianyu Zuo

\thanks{This work was supported in part by the National Natural Science Foundation of China (Grant 62202141), in part by Henan Province Science and Technology Research Project (Grant 232102240020), in part by University Young Key Teacher of Henan Province (Grant 2020GGJS027) and in part by Henan Provincial Education Science Planning Major Bidding Project "Research on Education Public Opinion under the Background of Informatization" (Grant 2018-JKGHZDZB-17). (Corresponding author: Lei Zhang.)}
	
\thanks{Yakun Niu, Lei Tan, Lei Zhang, and Xianyu Zuo are with Henan Key Laboratory of Big Data Analysis and Processing, Henan University, Kaifeng 475000, China, and also with School of Computer and Information Engineering, Henan University, Kaifeng 475000, China. (e-mail: \{ykniu, tanlei, zhanglei, xianyu\_zuo\}@henu.edu.cn).}
}
\markboth{Journal of \LaTeX\ Class Files,~Vol.~14, No.~8, August~2021}%
{Shell \MakeLowercase{\textit{et al.}}: A Sample Article Using IEEEtran.cls for IEEE Journals}

\maketitle

\begin{abstract}
Image operation chain detection techniques have gained increasing attention recently in the field of multimedia forensics. However, existing detection methods suffer from the generalization problem. Moreover, the channel correlation of color images that provides additional forensic evidence is often ignored. To solve these issues, in this article, we propose a novel two-stream multi-channels fusion networks for color image operation chain detection in which the spatial artifact stream and the noise residual stream are explored in a complementary manner. Specifically, we first propose a novel deep residual architecture without pooling in the spatial artifact stream for learning the global features representation of multi-channel correlation. Then, a set of filters is designed to aggregate the correlation information of multi-channels while capturing the low-level features in the noise residual stream. Subsequently, the high-level features are extracted by the deep residual model. Finally, features from the two streams are fed into a fusion module, to effectively learn richer discriminative representations of the operation chain. Extensive experiments show that the proposed method achieves state-of-the-art generalization ability while maintaining robustness to JPEG compression. The source code used in these experiments will be released at https://github.com/LeiTan-98/TMFNet.
\end{abstract}

\begin{IEEEkeywords}
Image forensics, multi-channels fusion, operation chain detection, convolutional neural network.
\end{IEEEkeywords}

\section{Introduction}
\IEEEPARstart{D}{igital} images have become an important information carrier in people's daily lives, widely used in fields such as news reporting, judicial appraisal, identity recognition, and online payment. However, with the vigorous development of social media and the widespread popularity of smart devices, people can forge digital image content anytime and anywhere, posing a huge threat to the authenticity of digital images. In order to verify the authenticity of the image and trace its processing history, many detection methods have emerged to analyze what operations the image has undergone, such as resampling\cite{popescu2005exposing,zhang2020robustness,vazquez2017random}, median filtering\cite{chen2015median,tang2018median,yuan2011blind,chen2013blind}, contrast enhancement\cite{stamm2010forensic,cao2014contrast,de2015second}, sharpening\cite{cao2011unsharp,ding2014novel,ding2014edge}, JPEG compression\cite{fan2003identification,luo2010jpeg,wang2021detecting}, and general-purpose operation\cite{qiu2014universal,fridrich2012rich,li2016identification}, etc.

\begin{figure}[t]
	\centering
	
	{
		\includegraphics[width=3in]{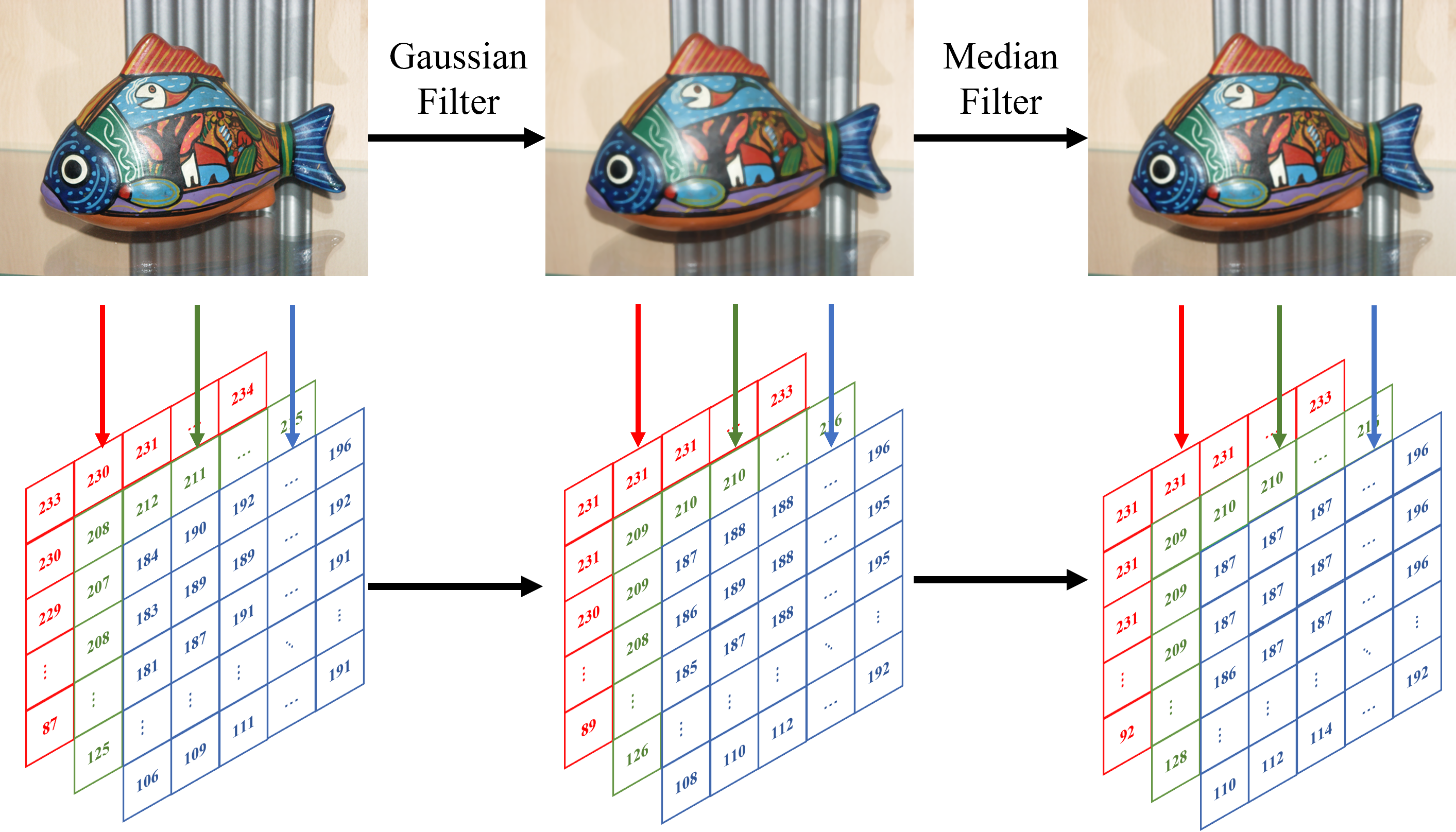}
	}
	
	\caption{Schematic diagram of the visual changes and pixel value changes of the image after two operations.}
	\label{fig:subfig_1}
	\vspace{-1em}
\end{figure}

Image operation chain refers to a series of processing operations on digital images, aiming to create more natural and realistic forged images or hide traces of operation. As shown in Fig. \ref{fig:subfig_1}, an original image has successively undergone Gaussian blur and median filtering operations by left to right. Clearly, it is difficult for us to distinguish between the three images from the naked eye, since human vision cannot directly observe the subtle changes in image content caused by the operation chain. However, it can be seen from the pixel value maps that the pixel values of the corresponding channels in the three images are quite different. Therefore, the potential patterns of the pixel changes caused by sequential operations can be as the clue for operation chain detection. Besides, the second operation overwrites the pixel values in three channels of the image obtained through the first one to some extent. Meaning that the traces left by the previous operation could be weakened or even erased by the following one. As a result, methods designed for identifying specific operations may fail. Specially, with the continuous development of image processing software and tools, the complexity and invisibility of operation chains are becoming increasingly high, making image operation chain detection a very challenging task.

Recent years, researchers have widely applied convolutional neural network (CNN) to image forensics, such as copy-move\cite{li2022image,cozzolino2015efficient} and splicing forgery\cite{rao2020deep,krishnamoorthy2022splicing} detection, etc. However, there are still few researches on the use of deep learning in image operation chain detection. Amerini $et$ $al$.\cite{amerini2017localization} designed a multi domain CNN for detecting dual JPEG compression, given a to-be checked image, is able to reliably localize the possible forged areas by analyzing the presence of single or double JPEG compressed areas. Bayar $et$ $al$.\cite{bayar2018constrained} proposed a novel CNN with the Constrained Convolutional Layer (CCL) for detecting image operator chains. CCL can jointly suppress the content of the image and adaptively learn image operation detection features directly from the data. Barni $et$ $al$.\cite{barni2018cnn} proposed a CNN model for detecting contrast adjustment in JPEG post-processing, which is robust to JPEG compression. The used system relies on a patch-based CNN, trained to distinguish pristine images from contrast adjusted images, for some selected adjustment operators of different nature. Recently, Chen $et$ $al$.\cite{chen2023identification} proposed a feature decoupling based image operation chain recognition method, inspired by blind signal separation, which can eliminate superimposed processing artifacts and derive a set of decoupling operation features representing tampering operations in the operator chain. In\cite{you2021transformer}, a machine translation based  framework is proposed for image operation chain detection, in which the operation chain is considered as a sentence in the target language, and each operation in the chain is represented by a word. The authors first transforms the original image into a potential source language space, where the sentences can appropriately describe the fingerprints left by different operation chains. Then, by translating the corresponding sentences in the source language step by step into sentences in the target language, the operation chain is decoded. Their research has made gratifying progress for image operation chain detection.

However, the above image operation chain detection methods fail to fully take into account the richer operation information of color images. Based on the work in the literature \cite{you2021transformer}, Li $et$ $al$. \cite{you2022transformer} proposed a chain reversal modeling module that enables more accurate detection of long chains of operations. Liao $et$ $al$.\cite{liao2020robust} proposed a robust detection of image operator chains based on dual-stream CNNs. A data-driven sequential detection framework is used to automatically learn and obtain operation fingerprints. Various carefully designed preprocessing operations have been cleverly used for different image operator chains, and transfer learning strategies have been used to improve performance. Verde $et$ $al$.\cite{verde2023multi} employed a supervised framework containing a cascade of backtracking blocks, for reconstructing image sharing chains on social media platforms. However, they all require human intervention. In other words, the input needs to be preprocessed. It means that their model cannot effectively learn potential features resulting in poor generalization ability.

 To address these issues, we design a novel two-stream multi-channels fusion networks in which the spatial artifact stream and the noise residual stream are explored in a unified framework. Specifically, the global features extracted by multiple residual blocks without pooling in the spatial stream and low-high levels noise residuals obtained in the noise stream are fused in a mutually reinforcing way. The contributions of the paper are mainly as follows:a

\begin{itemize}
	\item{An in-depth analysis of the channel correlation of the color image is conducted and it derives that the effect of different operation chains on channel correlation is obviously different. Based on this fact, we propose a novel two-stream multi-channels fusion networks, which fully mines the latent complementary relationships between the spatial artifact stream and the noise residual stream.}
	\item{We first propose a novel deep residual architecture without pooling in the spatial artifact stream for learning the global features representation of multi-channels correlation. In the noise residual stream, a set of filters is designed to aggregate the correlation information of multi-channels and then both low-level and high-level features are extracted by a subsequent deep network.}
	\item{Extensive experimental results show that the proposed method achieves very competitive generalization ability in cross-database, cross-resolution, and without prior while maintaining robustness to JPEG compression. Moreover, our approach also has satisfactory performance in detecting long operation chains or sharing chains on social platforms.}

\end{itemize}

The rest of the paper is organized as follows. Section II introduces the preliminary knowledge of the image operation chain briefly. Section III presents the proposed networks for color image operation chain detection. Section IV reports and analyzes the experimental results. Section V provides relevant discussions. Finally, a conclusion is made in Section VI.

\section{Preliminary}
Color images are the most commonly encountered image types in our daily lives. To express a color image, multiple color channels are usually used to form a color space together. Color space is a three-dimensional description of color vision, where each color can be represented by a point. When the image is manipulated, it is inevitable to break the channel correlation to some extent. Therefore, channel correlation can be an important clue for multimedia forensics.

The rich channel information of color images has been widely used for image steganography and forgery localization. Liao $et$ $al$.\cite{liao2020adaptive} utilized the correlation of the three RGB channels of color images to adjust the distortion cost to improve security. In \cite{guo2023exposing,zhou2018learning}, the authors found the color difference in color images between original and tampered regions are more pronounced than that in grayscale images. The detection accuracy and robustness can be greatly improved by utilizing the correlation between color channels. Inspired by\cite{liao2020adaptive,guo2023exposing,zhou2018learning}, we investigate color channels correlation for color image operation chain detection in this paper.

To analyze the influence of different operation chains on color channels, the Pearson correlation coefficient is employed to measure the channel correlation, which can be obtained by:
\begin{equation}
	\begin{gathered}
		R_{xy}=\frac{\left|cov\left(x,y\right)\right|}{\sqrt{D\left(x\right)}\sqrt{D\left(y\right)}},
	\end{gathered}
\end{equation}
where $R_{xy}$ is the correlation coefficient between channels $x$ and $y$. $cov(\cdot)$ and $D(\cdot)$ denote the covariance and the variance respectively. The greater the $R_{xy}$, the stronger the correlation.
	\begin{table}[h]
	\setlength{\tabcolsep}{3pt}
	\centering
	\caption{Correlation coefficients between different channels with different operation chains on RAISE dataset\cite{dang2015raise}.}
	\renewcommand\arraystretch{1.2}
	\begin{tabular}{|c|c|c|c|c|c|c|}
		\hline
		Correlation&  Channels & Original  & MF & GB & MF$\rightarrow$GB & GB$\rightarrow$MF\\
		\hline
		\multirow{3}*{Mean} & R/G & 0.9450 & 0.9278 & 0.9306 & 0.9014 &0.8957  \\
		\cline{2-7}
		& G/B & 0.8961 & 0.8895 & 0.8928 & 0.8633 & 0.8798  \\
		\cline{2-7}
		& R/B & 0.8108 & 0.8127 & 0.8069 & 0.8082 &0.7939  \\
		\hline
		\multirow{3}*{Variance} & R/G & 0.0131 & 0.0082 & 0.0060 & 0.0139 & 0.0194  \\
		\cline{2-7}
		& G/B & 0.0116 & 0.0097 & 0.0109 & 0.0133 & 0.0135  \\
		\cline{2-7}
		& R/B & 0.0280 & 0.0221 & 0.0183 & 0.0228 &0.0178  \\
		\hline
	\end{tabular}
	\label{tab:table_0}
	\vspace{-1em}
\end{table}

Table \ref{tab:table_0} shows the mean and variance of the correlation coefficients between RGB channels with median filtering  (MF), Gaussian blur (GB), and both the two operations in different order.  From Table \ref{tab:table_0}, we can see that the distribution of the correlation coefficients between color channels with different operation chains are quite different. It means that the channel correlation of color images can provided additional  forensics evidence for image operation chain detection.

\section{Proposed Method}
In this section, for the sake of clarity, we first define the problem of the color image operation chain. Next, we will provide a detailed description of the design concept and specific implementation details of the network architecture.
\subsection{Problem Formulation}

We first provide the definitions of all operations and their parameters in Table \ref{tab:table_1}. The operation chain is defined as an ordered sequence of these operations applied to an image. Assuming that we use three operations, e.g., $O_{MF}$, $O_{GB}$, and $O_{RS}$, to form a chain with a maximum length $N=2$, and each operation is used no more than once. There are $A^0_3+A^1_3+A^2_3=1+3+6=10$ possible operation chains:

\begin{equation}
	\begin{split}
		C_1 &= O_{AU} \text{: An unaltered image.}\\
		C_2 &= O_{MF} \text{: Altered by $O_{MF}$ only.}\\
		C_3 &= O_{GB} \text{: Altered by $O_{GB}$ only.}\\
		C_4 &= O_{RS} \text{: Altered by $O_{RS}$ only.}\\
		C_5 &= O_{MF} \rightarrow O_{GB} \text{: Altered by $O_{MF}$ then by $O_{GB}$.}\\
		C_6 &= O_{MF} \rightarrow O_{RS} \text{: Altered by $O_{MF}$ then by $O_{RS}$.}\\
		C_7 &= O_{RS} \rightarrow O_{MF} \text{: Altered by $O_{RS}$ then by $O_{MF}$.}\\
		C_8 &= O_{RS} \rightarrow O_{GB} \text{: Altered by $O_{RS}$ then by $O_{GB}$.}\\
		C_9 &= O_{GB} \rightarrow O_{RS} \text{: Altered by $O_{GB}$ then by $O_{RS}$.}\\
		C_{10} &= O_{GB} \rightarrow O_{MF} \text{: Altered by $O_{GB}$ then by $O_{MF}$.}\\
		\end{split}
\end{equation}
Here, the symbol $A_m^n$ is a mathematical concept that represents taking $n$ out of $m$ different operations and arranging them in a certain order. e.g., $A^1_3$ indicates that only one of the three operations is involved. Note that $C_1$ denotes the chain of length 0, i.e., no operation has been performed on the image. Our goal is to identify the correct operation chain applicable to the image from the set of candidate operation in Eq. (2). In this way, we transform the operation chain detection problem into a classification problem.

\begin{figure*}[htp]
	\centering
	\includegraphics[width=6in]{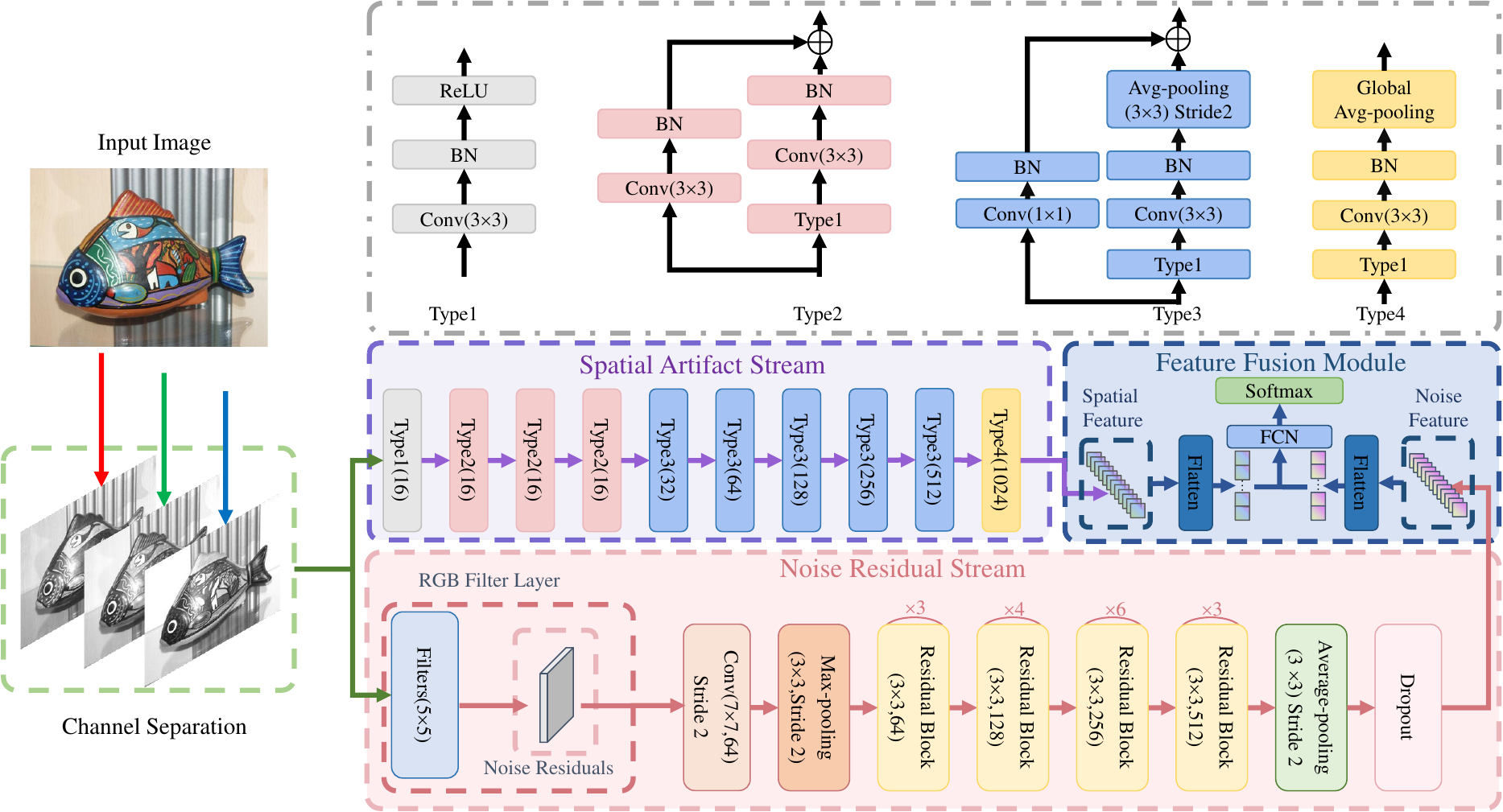}
	\caption{Overview of the network architecture diagram of the proposed TMFNet framework.}
	\label{fig:subfig_4}
	\vspace{-1em}
\end{figure*}

\begin{table}[t]
	\setlength{\tabcolsep}{1.8pt}
	\caption{Operation Dictionary\label{tab:table_1}}
	\centering
	\renewcommand\arraystretch{1.5}
	\begin{tabular}{|c|c|c|}
		\hline
		Symbol & Operation & Parameter\\
		\hline
		$O_{AU}$ & Original& -\\
		\hline
		$O_{MF}$ & Median Filter (MF)& kernel size = 3, 5\\
		\hline
		$O_{GB}$&Gaussian Blur (GB)& kernel size = 5 $\sigma$ = 0.7, 1.0, 1.1\\
		\hline
		$O_{JPEG}$ & JPEG Compression (JPEG)& QF = 70 , 75 , 80 , 85 , 90\\
		\hline
		$O_{RS}$ & Resampling (RS)& Scaling = 1.2 , 1.5\\
		\hline
		$O_{USM}$ & Unsharp Masking (USM)& $\lambda$ = 1\\
		\hline
		$O_{HE}$ & Histogram Equalization (HE)& -\\
		\hline
		\multirow{2}*{$O_{AWGN}$} &Additive White Gaussian&\multirow{2}*{$\sigma=2$} \\ & Noise (AWGN) & \\
		\hline
	\end{tabular}
	\vspace{-1.5em}
\end{table}
\subsection{The Architecture of TMFNet}

The architecture of the TMFNet network, as shown in Fig. \ref{fig:subfig_4}, consists of two streams. One is the spatial artifact stream, and the other one is the noise residual stream. The spatial artifact stream receives color images as input and extracts operation artifact features through the CNN layer. These features express the global operation artifact information presented in the image. The noise residual stream preprocesses the input image through the RGB filters, and then uses a CNN layer to extract the local noise residual features. Features extracted from these two streams are processed through a fully connected layer and then concatenated together. Finally, the input images are classified into specific categories through the Softmax layer. In the following, we will describe these two streams in detail.

\subsubsection{Spatial Artifact Stream}

In the spatial artifact stream part, we mainly extract the global operation artifact features of the image. Although the image texture will affect such feature extraction, the fused features of operation fingerprint and texture would be of great help to our forensic work. In order to extract the spatial features of the image as much as possible, we reduce the use of the pooling operation in the spatial artifact stream, because it will destroy the information distribution of spatial features. In addition, by cleverly designing the model structure in the feature extraction phase, we achieve the global receptive field with only a small number of feature extraction blocks, and in this way successfully capture the global features.

We adopted the design philosophy of residual networks to construct the spatial artifacts stream, which consists of four parts. The first part comprises only one convolutional layer, using 16 sets of $3\times3$ convolution kernels to convolve the input 3-channel RGB image $X\in \mathbb{R}^{3\times H\times W}$. After convolution, Batch Normalization (BN) is applied for normalization, followed by passing the ReLU activation function, generating 16 feature maps to represent the shallow features of image operation artifacts:
\vspace{-1em}
\begin{equation}
	\begin{split}
		x^1_i=& f(F(\textbf{W}_i,X)),
	\end{split}
\end{equation}
where $x_i^1$ denotes the $i$-th output feature map of first layer, $\textbf{W}_i$ denotes the $i$-th weight matrix, $F(\cdot)$ denotes the linear mapping and $f(\cdot)$ denotes the activation function.

Subsequently, the extracted feature maps from the first part are forwarded to the second part to extract deeper operation features. The second part consists of three residual blocks. Each residual block first employs 16 convolutional operations with $3\times3$ convolution kernels on the input feature map, and then concatenates the resulting 16 feature maps with the input feature map, achieving the fusion of deep and shallow features related to image operation artifacts. The operation process can be written as:
\vspace{-0.5em}
\begin{equation}
	\begin{split}
		x^2_i =& F(\textbf{W}_i,f(F(\textbf{W},x^1)))+ F(\textbf{W}_i,x^1),
	\end{split}
\end{equation}
where $x^2_i$ denotes the $i$-th output feature map of this layer, $\textbf{W}$ denotes the set of weight matrices for the layer.

The first two parts are crucial for extracting global operation artifacts in the entire network. Due to the fact that average pooling acts as a low-pass filter, enhancing content and suppressing noise by averaging changes in nearby embeddings, we disable average pooling in the first two parts.
\begin{figure}[htp]
	\centering
	\includegraphics[width=1.65in]{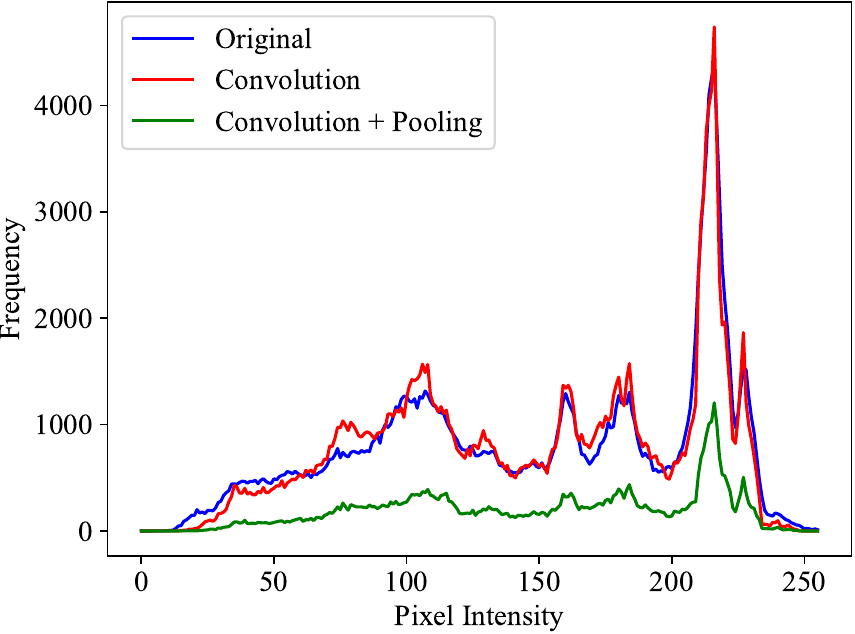}
	\caption{The distribution of pixel intensity in R-channel. }
	\label{fig:subfig_5}
	\vspace{-1em}
\end{figure}

To illustrate the above conclusion, we conduct experiments on RAISE database. Fig. \ref{fig:subfig_5} shows the distributions of pixel intensity in the R channel of original images, convolutional images, and convolutional followed by pooling images. It can be clearly seen that convolution has little effect on image pixels while pooling significantly destroy the distribution of pixels. Therefore, we disable the pooling operation in the feature extraction phase.

The third part includes five layers, which utilize residual connections to further obtain deeper operation features while reducing information redundancy. We first perform convolution with $3\times 3$ kernel size and normalization on the input feature map, followed by a pooling operation with a kernel size of $3\times 3$ and a stride of 2. Simultaneously, the input feature map undergoes $1\times 1$ convolution with a stride of 2, and the outputs of both operations are concatenated. This helps reduce parameters for each feature dimension and prevents overfitting. The above process in each layer can be given by:
\begin{equation}
	\begin{split}
		x^3_i =& Avg(F(\textbf{W}_i,f(F(\textbf{W},x^2))))+ F(\textbf{W}_i,x^2),
	\end{split}
\end{equation}
where $Avg(\cdot)$ is the average pooling layer.

In the fourth stage, after the convolution operation with a $3\times 3$ kernel, normalization is conducted. Subsequently, global average pooling is employed to obtain the feature $x_i^4\in\mathbb{R}^{1024\times 1\times 1}$:
\vspace{-0.5em}
\begin{equation}
	\begin{split}
		x^4_i =& G(F(\textbf{W}_i,f(F(\textbf{W},x^3)))),
	\end{split}
\end{equation}
where, $G(\cdot)$ is the global average pooling layer.


In \cite{boroumand2018deep}, although SRNet also utilizes a deep residual structure \cite{he2016deep} that is very similar to ours for a different forensic task, namely steganalysis, there are significant differences between our spatial artifact stream and SRNet. Unlike SRNet which first extracts high-dimensional features and then reduces their dimensions, our approach directly extracts low-dimensional feature maps using convolution, thereby preserving the global operation artifacts to the greatest extent. Meanwhile, rather than using simple residual concatenation, we combine convolution and batch normalization (Conv+BN) to learn global feature representations and enhance the model's receptive field. Furthermore, to obtain deeper features while reducing information redundancy, our model reduces the feature maps into a higher-dimensional vector than SRNet, improving the model's feature representation capability. Last but not least one, we use fewer feature extraction layers and more feature fusion layers, which reduces the number of parameters while maintaining performance and allows the model to handle high-resolution input images effectively.

\begin{figure}[t]
	\centering
	\includegraphics[width=2.8in]{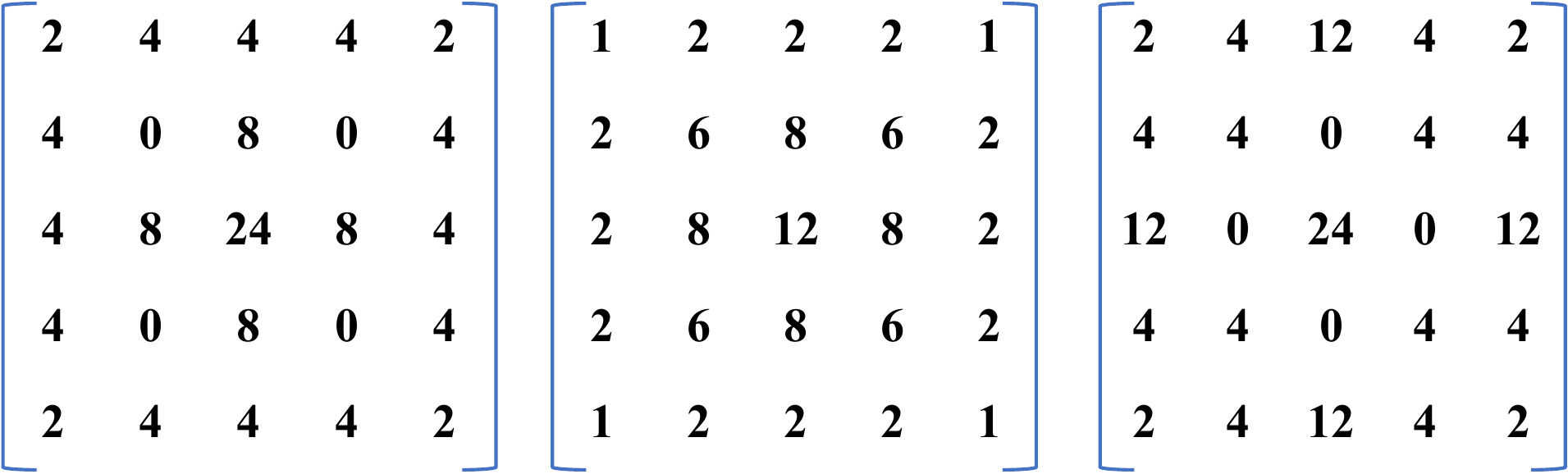}
	\caption{The proposed three RGB filters for image texture suppression.}
	\label{fig:subfig_11}
	\vspace{-1em}
\end{figure}

\begin{figure}[t]
	\centering
	\includegraphics[width=1.8in, keepaspectratio]{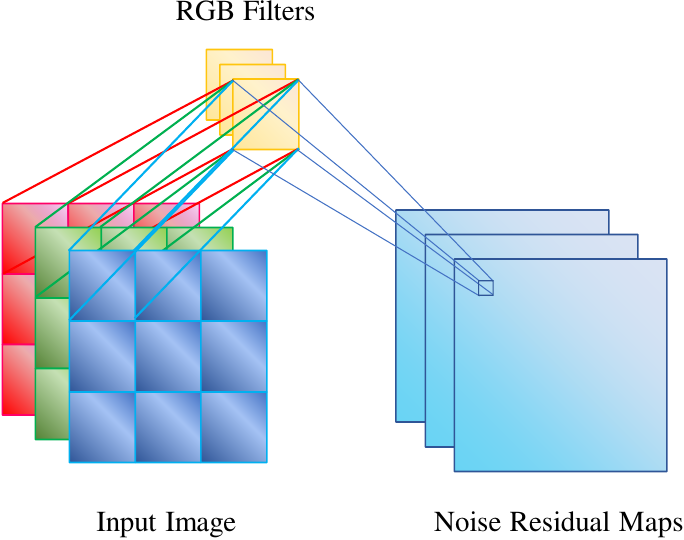}
	\caption{Schematic diagram of the noise residual maps extraction with the RGB filters.}
	\label{fig:subfig_2}
	\vspace{-1.5em}
\end{figure}

\subsubsection{Noise Residual Stream}

The spatial artifact stream is more concerned with the global operation artifact of the image in RGB space. To learn the local noise residual evidence of the operation chain, we design the noise residual stream that provides additional forensic information. In contrast to the spatial artifact stream, the noise residual stream pays more attention to noise rather than image content. 

To better capture the operation noise features, previous researchers have done some attempts to suppress image texture. In \cite{fridrich2012rich}, the Spatial Rich Model (SRM) was proposed, which can suppress texture information of images while highlighting noise information in the image. However, we have found through a series of experimental studies that the SRM model has a good texture suppression effect on grayscale images, but the texture suppression effect on color images is unsatisfactory. Bayar $et$ $al$.\cite{bayar2018constrained} proposed a self-learning constrained CNN method for image manipulation detection. The weights of CCL are randomly initialized before the model training and then they are continuously updated in subsequent training to better fit specific training tasks. The constrained CNN is a universal network architecture that can self-learn different weights based on different tasks and has been widely applied in areas such as image steganalysis and image tampering detection. However, due to the random selection of the initial values, CCL cannot suppress the content well for some images, specially for those with smooth background. 

In this paper, to efficiently suppress image texture and obtain the noise information, hand-designed RGB filters are proposed in the noise residual stream. The design principles are mainly reflected in two aspects. Firstly, we make the filter weights obey the Gaussian distribution, which is defined as:
\begin{equation}
		\begin{gathered}
		g_j(x,y)= \frac {1}{2\pi \sigma^ {2}}$exp$({-\frac {x^ {2}+y^ {2}}{2\sigma^ {2}}}),
		\end{gathered}
\end{equation}
where $\sigma$ is the standard deviation. It assigns weights to pixels based on their distance from the central pixel in the filter's neighborhood. This process effectively blurs the image, smoothing out sharp edges and reducing high-frequency components, thereby achieving the effect of noise reduction.  However, the Gaussian function depends on the standard deviation $\sigma$ which may lead to an unsatisfactory filtering effect. Inspired by SRM, we then constrain the filter weights based on the pixel predictor:	
\begin{equation}
		\begin{gathered}
			S_{i,j}=Pred(N_{i,j})+cX_{i,j},
		\end{gathered}
\end{equation}
where $N_{i,j}$ is a set of neighboring pixels of $X_{i,j}$, $c\in \mathbb{N^+}$ is the weight coefficient , and $Pred(\cdot)$ is the predictor. By using Eq. (8) to constrain Eq. (7), the proposed RGB filters are able to combine the advantages of SRM and Gaussian filtering. That is, they not only suppress the image texture but also enrich the extraction of operation noise features.

	\begin{table}[t]
	\centering
	\caption{Effect of different filters on neighborhood pixel correlation.}
	\renewcommand\arraystretch{1.5}
	\begin{tabular}{|c|c|c|c|c|}
		\hline
		Filter &Channel & Horizontal & Vertical & Diagonal     \\
		\hline
		\multirow{3}*{Original}&R & 0.9786 & 0.9900 & 0.9701 \\
		\cline{2-5}
		&G & 0.9683 & 0.9811 & 0.9535 \\
		\cline{2-5}
		&B & 0.9278 & 0.9594 & 0.9192 \\
		\hline
		\multirow{3}*{SRM\cite{fridrich2012rich}}&R & 0.3872 & 0.3223 & 0.2065\\
		\cline{2-5}
		&G & 0.5253 & 0.5013 & 0.3883\\
		\cline{2-5}
		&B & 0.6076 & 0.7193 & 0.5874\\
		\hline
		\multirow{3}*{CCL\cite{bayar2018constrained}}&R & 0.1660 & 0.2548 & 0.1954\\
		\cline{2-5}
		&G & 0.1611 & 0.2938 & 0.1847\\
		\cline{2-5}
		&B & 0.3129 & 0.3365 & 0.2901\\
		\hline
		\multirow{3}*{RGB}&R & 0.0107 & 0.0207 & 0.0317\\
		\cline{2-5}
		&G & 0.0078 & 0.0090 & 0.0105\\
		\cline{2-5}
		&B & 0.0122 & 0.0260 & 0.0085\\
		\hline
	\end{tabular}
	\label{tab:table_14}
	\vspace{-1em}
\end{table}

Based on Eqs. (7) and (8), we built three RGB filters to suppress image content, as shown in Fig. 4. They are constructed as locally supported linear filters that are combined to increase their output diversity. For ease of understanding, we can consider predicting each filter based on the pixel coefficients on the vertical, horizontal, and diagonal. For example, in the first and third filters, the center weights $X_{i,j}$ are set based on the sum of their elements in the horizontal and vertical directions, while in the second filter, the center weights $X_{i,j}$ are predicted as the sum of local elements in adjacent diagonal directions.

	The fusion of the three channels of the color image with the RGB filters can be  formulated as:
	\begin{equation}
		M_j(h,w)=\sum _{i=1}^{3}\sum _{x=1}^5\sum _{y=1}^5f_i(h,w)g_j(h-x,w-y),
	\end{equation}
	where $M_j$ represents the filtered result, $h\in[1,\dots,H]$ and $w\in[1,\dots,W] $ are coordinate indices, $f_i$ denotes the $i$-th color channel of the input image, and $g_j$ is the $j$-th RGB filter. Then, by subtraction of the original image and the filtered image, we can obtain the noise residual map. For ease of use in neural networks, we replicate and reassemble the filter to obtain a four-dimensional filter. Its working principle is to add the filtered information from each channel to aggregate information from the three channels, thereby suppressing the texture of the image. For ease of understanding, we have illustrated the working process of the filter in Fig. \ref{fig:subfig_2}. We can see that each filtered pixel represents the features of the $ 5\times 5 $ pixel block in the three channels of the original image. This means that the filter successfully aggregates three-channel information to suppress image textures and obtain low-level operation features. The pixel correlations in horizontal, vertical, and diagonal directions are utilized to evaluate the quality of texture suppression of different filters. Note that,  pixel correlation is calculated by Pearson correlation coefficient which is the quotient of the product of the covariance and the standard deviation between two pixels. The correlation coefficient of neighboring pixels belong to [0, 1] and the larger the value, the more relevant it is.
	
\begin{figure}[t]
	\centering
	\begin{minipage}{0.22\linewidth}
		\centerline{\includegraphics[width=0.7in,height=0.5in]{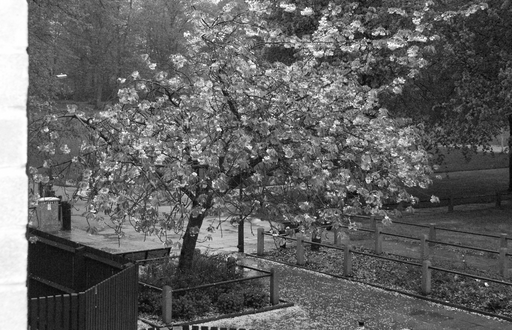}}
		\vspace{0.5em}
	\end{minipage}
	\begin{minipage}{0.22\linewidth}
		\centerline{\includegraphics[width=0.7in,height=0.5in]{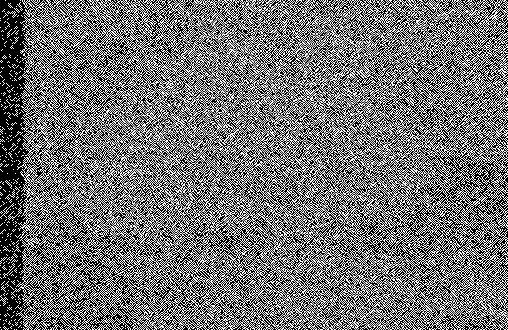}}
		\vspace{0.5em}
	\end{minipage}
	\begin{minipage}{0.22\linewidth}
		\centerline{\includegraphics[width=0.7in,height=0.5in]{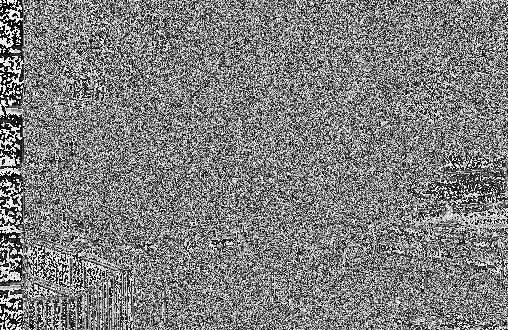}}
		\vspace{0.5em}
	\end{minipage}
	\begin{minipage}{0.22\linewidth}
		\centerline{\includegraphics[width=0.7in,height=0.5in]{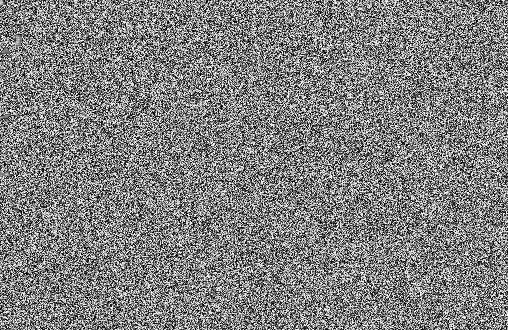}}
		\vspace{0.5em}
	\end{minipage}\\
	\begin{minipage}{0.22\linewidth}
		\centerline{\includegraphics[width=0.7in,height=0.5in]{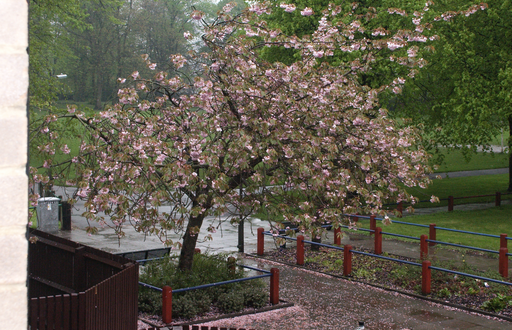}}
		\vspace{0.5em}
	\end{minipage}
	\begin{minipage}{0.22\linewidth}
		\centerline{\includegraphics[width=0.7in,height=0.5in]{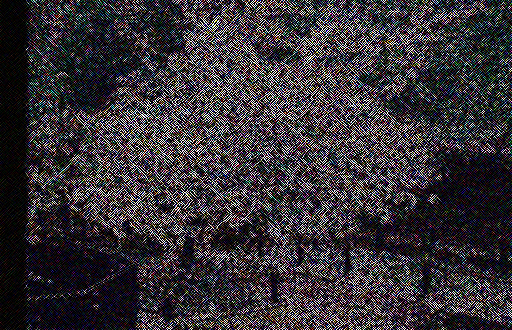}}
		\vspace{0.5em}
	\end{minipage}
	\begin{minipage}{0.22\linewidth}
		\centerline{\includegraphics[width=0.7in,height=0.5in]{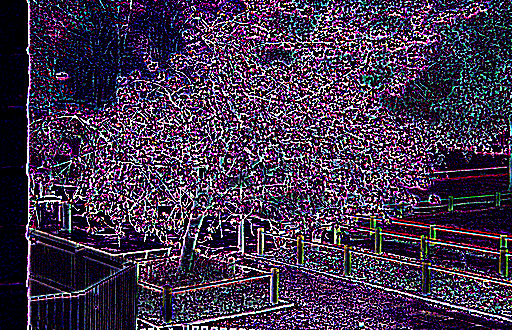}}
		\vspace{0.5em}
	\end{minipage}
	\begin{minipage}{0.22\linewidth}
		\centerline{\includegraphics[width=0.7in,height=0.5in]{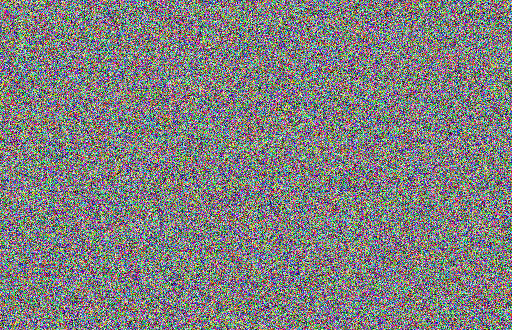}}
		\vspace{0.5em}
	\end{minipage}\\
	\begin{minipage}{0.22\linewidth}
		\centerline{\includegraphics[width=0.7in,height=0.5in]{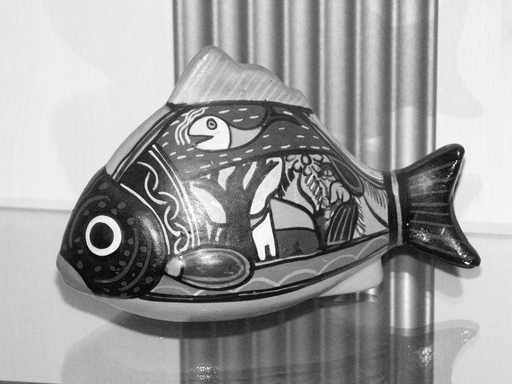}}
		\vspace{0.5em}
	\end{minipage}
	\begin{minipage}{0.22\linewidth}
		\centerline{\includegraphics[width=0.7in,height=0.5in]{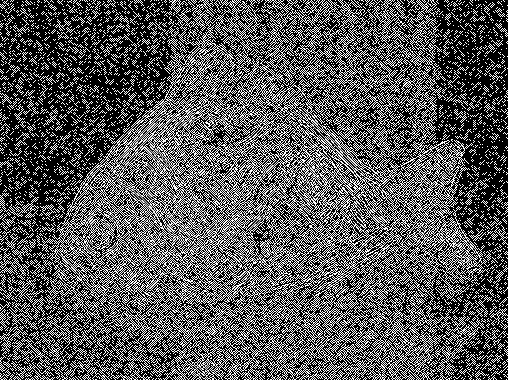}}
		\vspace{0.5em}
	\end{minipage}
	\begin{minipage}{0.22\linewidth}
		\centerline{\includegraphics[width=0.7in,height=0.5in]{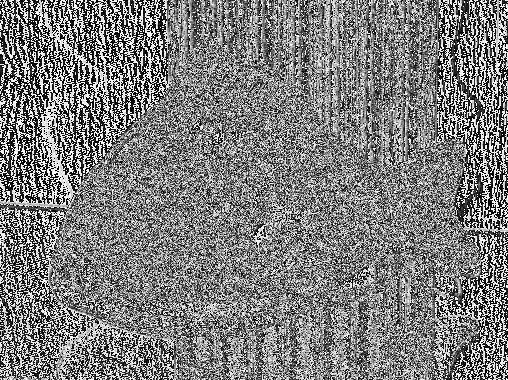}}
		\vspace{0.5em}
	\end{minipage}
	\begin{minipage}{0.22\linewidth}
		\centerline{\includegraphics[width=0.7in,height=0.5in]{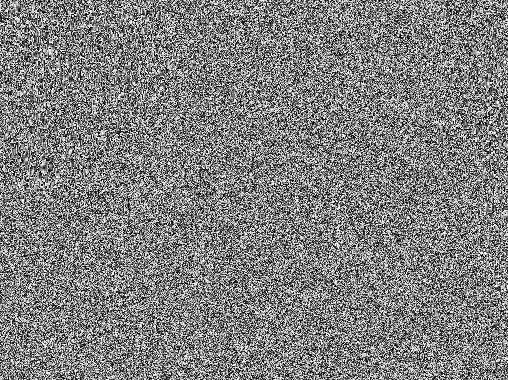}}
		\vspace{0.5em}
	\end{minipage}\\
	\begin{minipage}{0.22\linewidth}
		\centerline{\includegraphics[width=0.7in,height=0.5in]{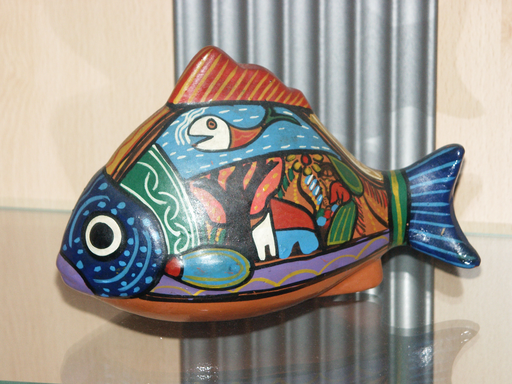}}
		\centerline{(a)}
	\end{minipage}
	\begin{minipage}{0.22\linewidth}
		\centerline{\includegraphics[width=0.7in,height=0.5in]{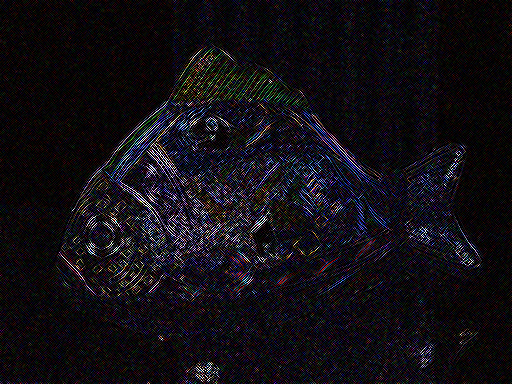}}
		\centerline{(b)}
	\end{minipage}
	\begin{minipage}{0.22\linewidth}
		\centerline{\includegraphics[width=0.7in,height=0.5in]{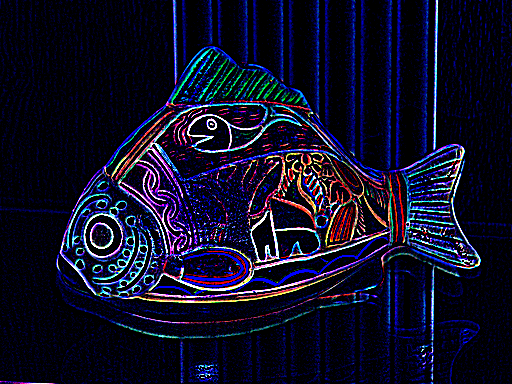}}
		\centerline{(c)}
	\end{minipage}
	\begin{minipage}{0.22\linewidth}
		\centerline{\includegraphics[width=0.7in,height=0.5in]{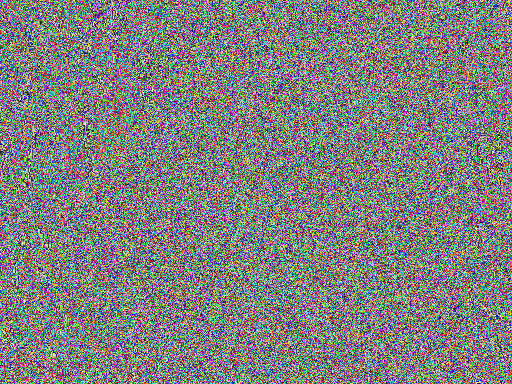}}
		\centerline{(d)}
	\end{minipage}\\
	\caption{The noise residuals obtained by (a) unfiltered image. (b) SRM filtered image. (c) CCL filtered image. (d) RGB filtered image.}
	\label{fig:subfig_3}
	\vspace{-1em}
\end{figure}

\begin{figure}[t]
	\centering
	\begin{minipage}{0.48\linewidth}
		\centerline{\includegraphics[width=1.7in,height=1.17in]{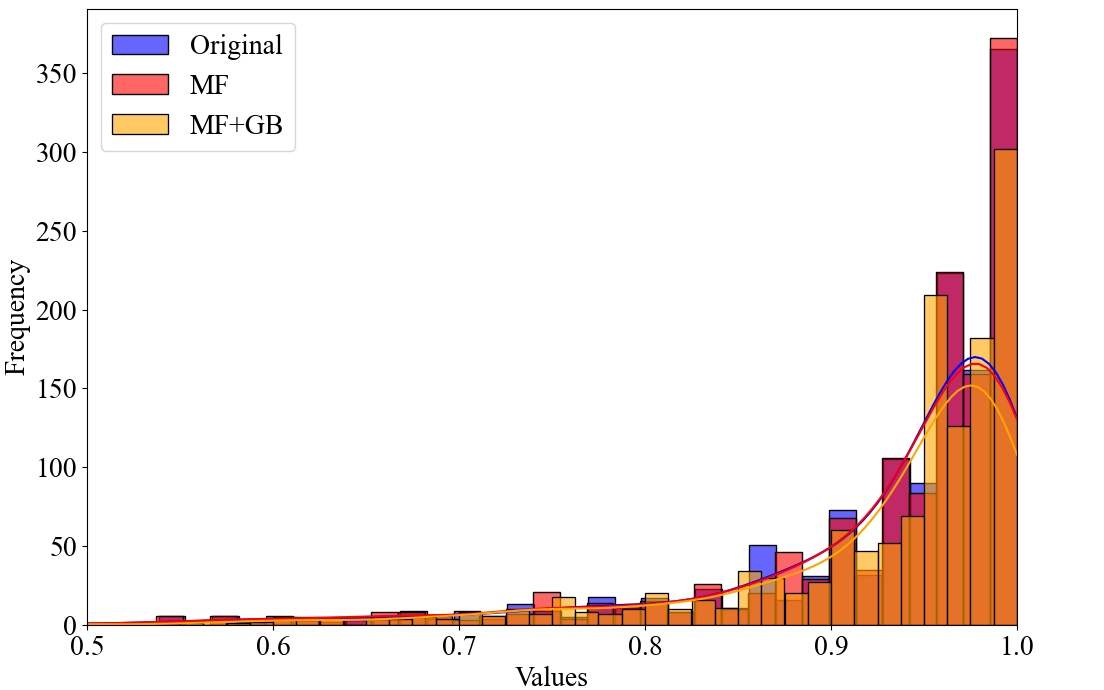}}
		\centerline{(a)}
	\end{minipage}
	\begin{minipage}{0.48\linewidth}
		\centerline{\includegraphics[width=1.7in,height=1.2in]{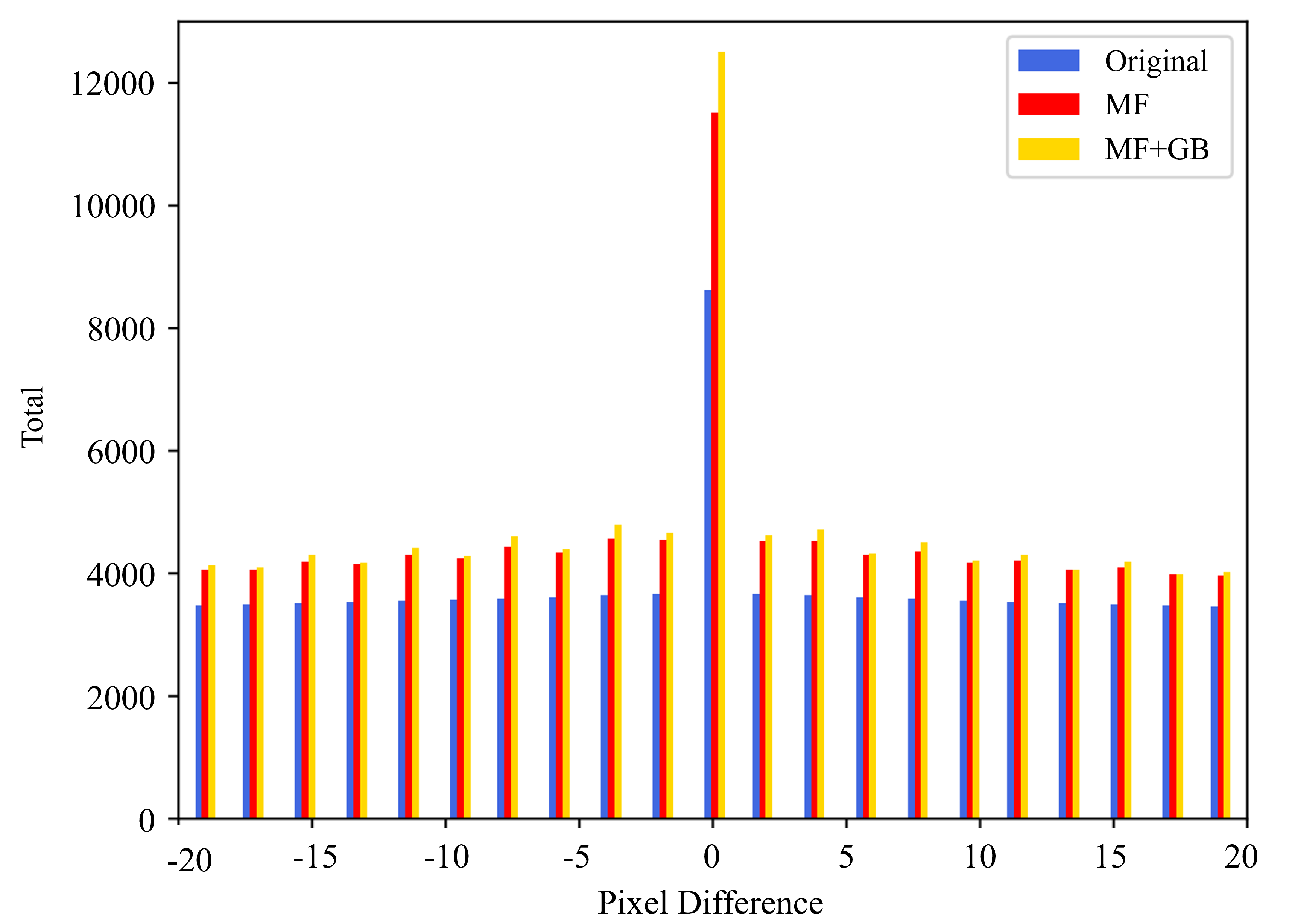}}
		\centerline{(b)}
	\end{minipage}\\
	\caption{(a) Statistical distribution of R/G correlation coefficient. (b) Neighborhood pixel difference of image filtered by different operations.}
	\label{fig:subfig_15}
	\vspace{-1.5em}
\end{figure}
Unlike the data-driven CCL that utilizes the weighted residuals between adjacent pixels and central pixel to update the filter weights, our model-based RGB filters employ weighted sum. In addition, our RGB filters also consider the inherent correlation between the three channels in color images, providing additional evidence for operation chain detection.
	
In order to illustrate the effectiveness of image texture suppression, Fig. \ref{fig:subfig_3} shows the noise residuals extracted by learnable filter CCL and hand-designed SRM and RGB filters. Clearly, SRM and CCL perform well in suppressing grayscale image textures, but rich texture information is retained in color images. On the contrary, our proposed RGB filters can effectively suppress the texture for both gray and color images. The reason is that RGB filters are able to fully aggregate the correlation information of multi-channels. Fig. \ref{fig:subfig_15} (a) shows the correlation coefficient distribution of different operation chains between R and G channels. We can see that the kernel density curves of the three chains are different. The curve of MF+GB is below that of MF, and so does MF and the original. Indicating that the correlation coefficients of long chains are smaller than those of short chains. Therefore, as concluded in Table \ref{tab:table_0}, the channel correlation can be used as a clue for operation chain detection. To further evaluate the effectiveness of the proposed RGB filters, Fig. \ref{fig:subfig_15} (b) shows the statistical distribution of neighborhood pixel difference of the noise residual obtained by RGB filters. Here, the blue, red and yellow bars correspond to original images, MF images and MF+GB images on RAISE database. One can see that the noise residual obtained by RGB filters can provide discriminative features for operation chain detection.

\begin{figure}[t]
	\centering
	\begin{minipage}{0.22\linewidth}
		\centerline{\includegraphics[width=0.7in,height=0.5in]{img/UCID_302}}
		\vspace{0.5em}
	\end{minipage}
	\begin{minipage}{0.22\linewidth}
		\centerline{\includegraphics[width=0.7in,height=0.5in]{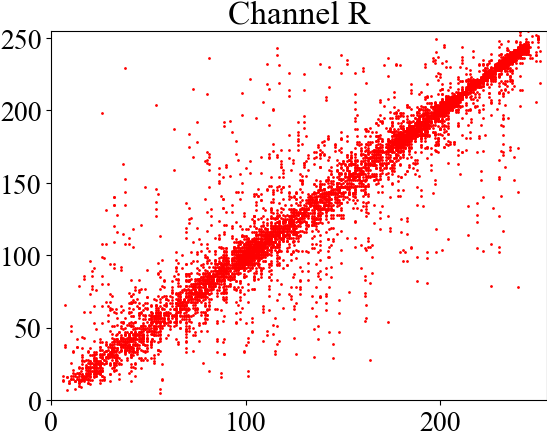}}
		\vspace{0.5em}
	\end{minipage}
	\begin{minipage}{0.22\linewidth}
		\centerline{\includegraphics[width=0.7in,height=0.5in]{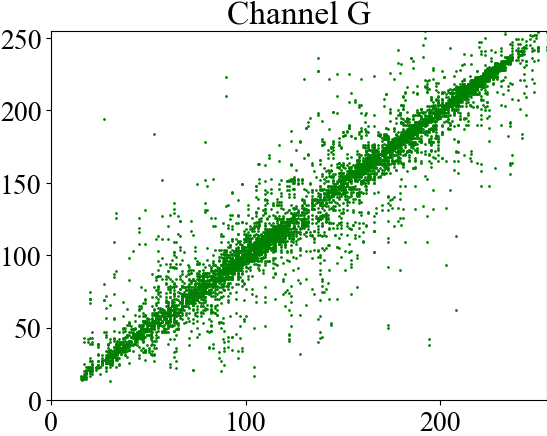}}
		\vspace{0.5em}
	\end{minipage}
	\begin{minipage}{0.22\linewidth}
		\centerline{\includegraphics[width=0.7in,height=0.5in]{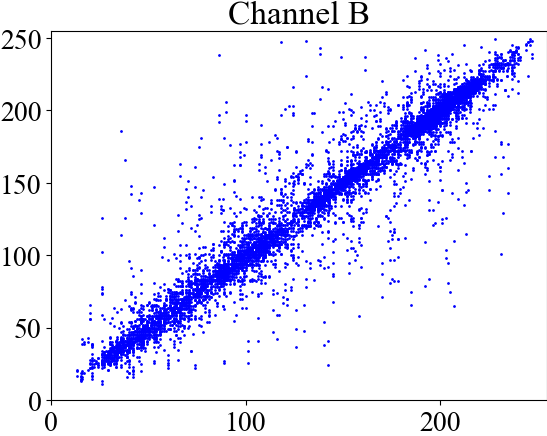}}
		\vspace{0.5em}
	\end{minipage}\\
	\begin{minipage}{0.22\linewidth}
		\centerline{\includegraphics[width=0.7in,height=0.5in]{img/SRM_image}}
		\vspace{0.5em}
	\end{minipage}
	\begin{minipage}{0.22\linewidth}
		\centerline{\includegraphics[width=0.7in,height=0.5in]{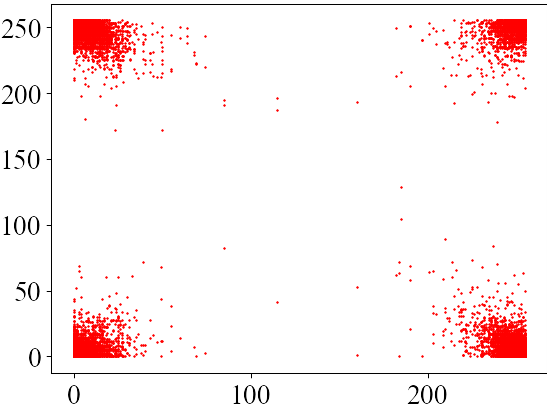}}
		\vspace{0.5em}
	\end{minipage}
	\begin{minipage}{0.22\linewidth}
		\centerline{\includegraphics[width=0.7in,height=0.5in]{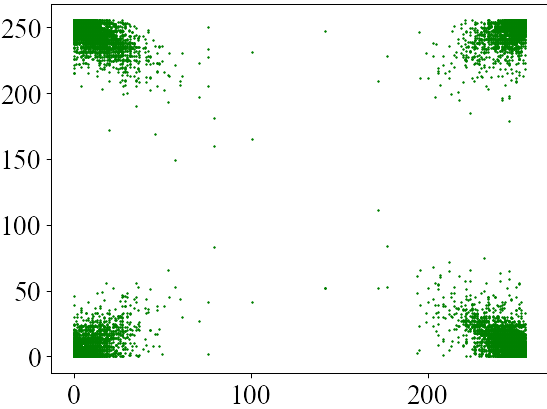}}
		\vspace{0.5em}
	\end{minipage}
	\begin{minipage}{0.22\linewidth}
		\centerline{\includegraphics[width=0.7in,height=0.5in]{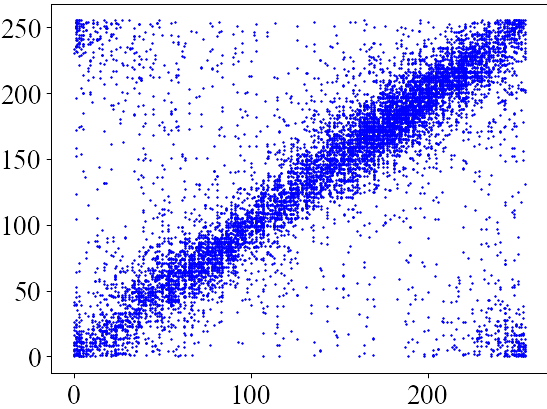}}
		\vspace{0.5em}
	\end{minipage}\\
	\begin{minipage}{0.22\linewidth}
		\centerline{\includegraphics[width=0.7in,height=0.5in]{img/constrain_image}}
		\vspace{0.5em}
	\end{minipage}
	\begin{minipage}{0.22\linewidth}
		\centerline{\includegraphics[width=0.7in,height=0.5in]{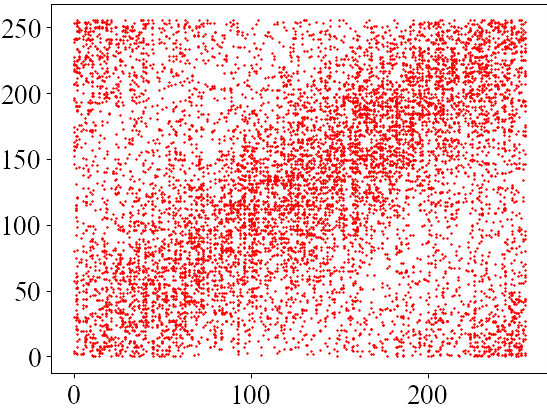}}
		\vspace{0.5em}
	\end{minipage}
	\begin{minipage}{0.22\linewidth}
		\centerline{\includegraphics[width=0.7in,height=0.5in]{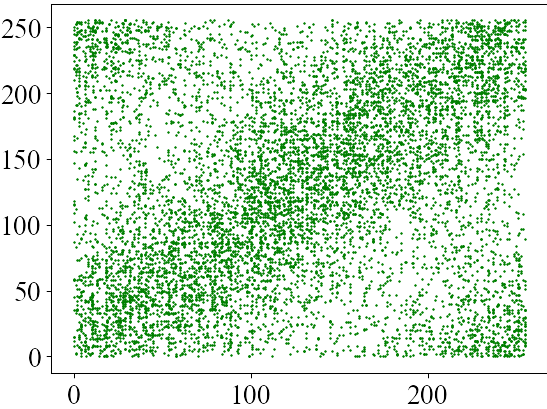}}
		\vspace{0.5em}
	\end{minipage}
	\begin{minipage}{0.22\linewidth}
		\centerline{\includegraphics[width=0.7in,height=0.5in]{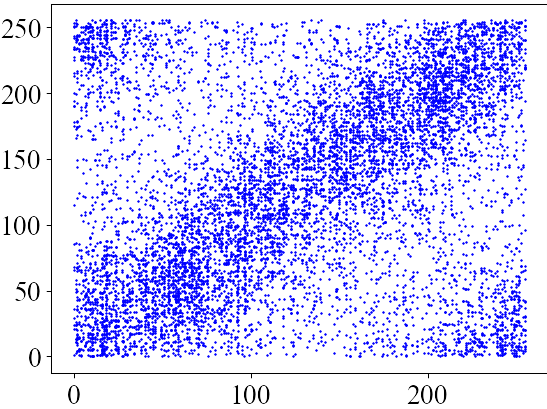}}
		\vspace{0.5em}
	\end{minipage}\\
	\begin{minipage}{0.22\linewidth}
		\centerline{\includegraphics[width=0.7in,height=0.5in]{img/our_image}}
		\centerline{(a)}
	\end{minipage}
	\begin{minipage}{0.22\linewidth}
		\centerline{\includegraphics[width=0.7in,height=0.5in]{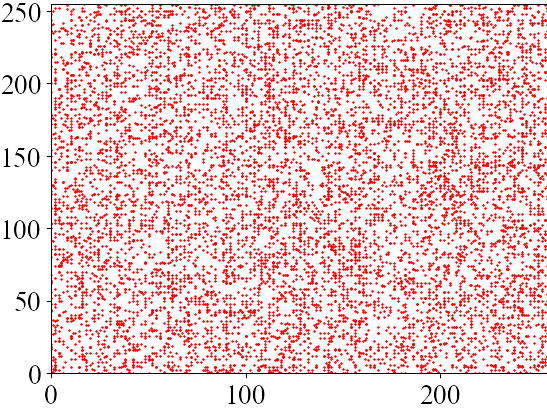}}
		\centerline{(b)}
	\end{minipage}
	\begin{minipage}{0.22\linewidth}
		\centerline{\includegraphics[width=0.7in,height=0.5in]{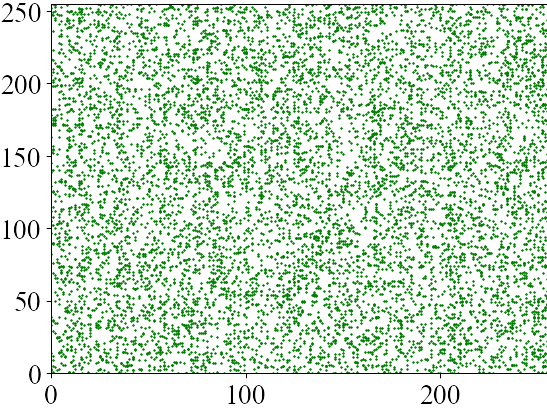}}
		\centerline{(c)}
	\end{minipage}
	\begin{minipage}{0.22\linewidth}
		\centerline{\includegraphics[width=0.7in,height=0.5in]{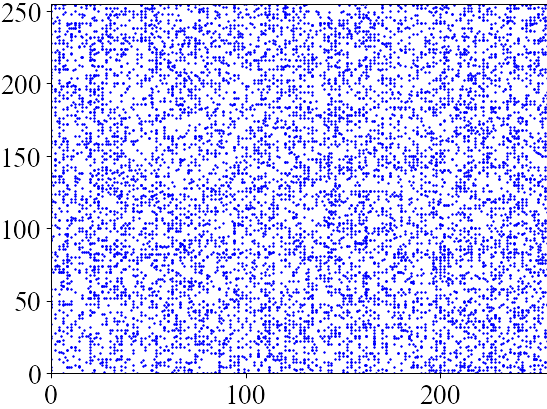}}
		\centerline{(d)}
	\end{minipage}\\
	\caption{From left to right: the noise residuals; the neighborhood pixel correlation coefficient on R channel; G channel and B channel. From top to bottom: the original image; the noise residual obtained by SRM; CCL and RGB filters.}
	\label{fig:subfig_14}
	\vspace{-1.5em}
\end{figure}

Furthermore, for more clearly prove the effectiveness of image texture suppression of the proposed RGB filters, we visualize the neighborhood pixel correlation on RGB channels in Fig. \ref{fig:subfig_14}. As can be seen from Fig. \ref{fig:subfig_14}, the neighborhood pixel correlation coefficients of the filtered images of SRM and CCL show clustering, which indicates that the filtered images still retain a certain degree of pixel correlation. On the contrary, the distribution of neighborhood pixel correlation is more scattered in the RGB-filtered image, indicating the complete destruction of neighborhood pixel correlation by the RGB filter, suggesting that it is effective in suppressing image texture. This is further confirmed by the results in Table \ref{tab:table_14}, where it can be found that the neighbor pixel correlation in horizontal, vertical, and diagonal directions tends to be 0. Indicating that the image texture is effectively suppressed.

Next, the low-level noise maps obtained from the texture suppression filter are sent to a network for high-level operation feature extraction. Here, ResNet50\cite{he2016deep} is adopted as the backbone of noise residual stream because of its efficient performance and concise network architecture. After feature extraction in the network, a certain proportion of neurons will be randomly discarded from the network during each iteration, with a dropout probability of 0.5. The purpose of neuron discarding is to prevent overfitting of the model and improve its generalization ability.

\subsubsection{Feature Fusion Module}

The global operation artifacts and the local noise residuals are integrated with the fusion module to achieve a more comprehensive and expressive feature representation. The feature fusion module adopts a fully connected concatenation mechanism, aiming to maximize the utilization of information from different streams and promote better understanding and expression of input data. In the first step, features extracted from two streams, denoted as $x_{s}$ and $x_{n}$, are transformed into one-dimensional feature maps, thereby achieving an end-to-end learning process for the model. Subsequently, the obtained mappings are concatenated together. The purpose is to merge information from both streams, creating an initial fused feature representation. Next, we feed the features into the fully connected layer to obtain a deeper representation of features. Finally, the obtained feature representation is input into the Softmax function, and the node with the highest probability is selected as our prediction target:
\begin{equation}
	\begin{split}
		P=Softmax(FCN(Cat(FN(x_{s}),FN(x_{n})))),
	\end{split}
\end{equation}
where $FN(\cdot)$ denotes the flattening the input features, $Cat(\cdot)$ denotes the splicing of the two input features, $FCN(\cdot)$ denotes the fully connected function, $P=[P_1,\dots,P_m]$ is the classification probability of the corresponding category and $m $ is the number of categories.

After the Softmax layer, we use cross-entropy loss for operation classification, which intuitively quantifies the prediction uncertainty of the model for each category, allowing the model to better learn the relationships between categories. Furthermore, it encourages the model to reduce the difference between predicted values and real labels, making it easier for the model to distinguish between different categories. In addition, the cross entropy loss function can accumulate the losses of various categories. This means that the model can simultaneously consider the classification accuracy of multiple categories and integrate the losses of each category as the overall optimization objective of the model. 

Thus, the total loss function can be expressed as:
\begin{equation}
	L_{total} = \lambda_1L_S + \lambda_2L_N
\end{equation}
where $L_S$ and $L_N$ denote the spatial artifact stream and the noise residual stream classification loss, respectively. $\lambda_1$ and $\lambda_2$ are the adjusting parameters. We experimentally find the best detection performance is achieved when $\lambda_1$ and $\lambda_2$ are equal compared to the other settings. For simplicity, we set $ \lambda_{1} = \lambda_{2} = 0.5$ in our experiment.

\section{Experimental Result}

Several state-of-the-art works \cite{bayar2018constrained,liao2020robust,you2022transformer} are selected as the baseline models for experimental comparisons. \cite{you2022transformer} is designed for grayscale images, while \cite{bayar2018constrained,liao2020robust} can be applied to both color and grayscale images. Firstly, we conduct comparative experiments between TMFNet and existing detection methods, including experiments on the accuracy of operation chain detection with prior knowledge and model robustness without prior knowledge. Next, we evaluate the generalization ability of TMFNet, including generalization evaluations for different resolutions and datasets. We also conduct the robustness of TMFNet against JPEG compression and then test its transfer learning ability. Finally, the performance of the proposed image filter is verified through comparative experiments, and the effectiveness of each component in the network is assessed through ablation experiments.

\subsection{Experimental Setup}

All our experiments are carried out on the NVIDIA Tesla V100 GPU. We implement our method using the PyTorch deep learning framework. In the experiment, we use SGD as the optimizer, the momentum value is fixed at 0.9, and the weight decay is 0.0005. We set 150 epochs for each experiment, with an initial learning rate of 0.01. The learning rate is updated every 30 epochs, adjusting it to 20\% of its previous value. Our batch size varies with the size of the training image. When training with $512\times512$ images, the batch size is set to 24, while for images with other resolutions, the batch size is set to 64. For transfer learning, we use the pre-trained model to initialize the CNN, dividing the initial learning rate by 10, and then proceed with CNN training. Meanwhile, the maximum iteration period and step size will be halved.
\subsection{Dataset}

\begin{table*}[t]
	\centering
	\caption{Confusion matrices for the operation chain identification using MISLnet \cite{bayar2018constrained}, Two-Stream CNN \cite{liao2020robust}, TransDetect \cite{you2022transformer} and our proposed TMFNet, where bold indicates the best accuracy, and underline indicates the second best.}
	\begin{tabular}{|c|c|c|c|c|c|c|c|c|c|c|}
		\hline
		\textbf{Operation}& {AU} & {MF} & {GB} & {RS} & {MF$\rightarrow$GB} & {GB$\rightarrow$MF} &{MF$\rightarrow$RS} & {RS$\rightarrow$MF} & {GB$\rightarrow$RS} & {RS$\rightarrow$GB} \\
		\hline
		&\multicolumn{10}{c|}{\textbf{MISLnet \cite{bayar2018constrained}: ACC = 94.40\% {{TFLOPs = 0.31 Params = 6.08M}}}} \\
		\hline
		
		{AU} &99.46\% &0.20\% &0.00\% &0.34\% &0.00\% &0.00\% &0.00\% &0.00\% &0.00\% &0.00\%\\
		\hline
		
		{MF} &0.00\% &88.46\% &0.00\% &0.00\% &0.00\% &0.49\% &1.81\% &8.96\% &0.28\% &0.00\% \\
		\hline
		
		{GB} &0.00\% &0.00\% &87.07\% &0.00\% &0.65\% &0.00\% &0.00\% &0.00\% &0.07\% &12.20\% \\
		\hline
		
		{RS} &0.40\% &0.00\% &0.07\% &99.12\% &0.00\% &0.00\% &0.27\% &0.13\% &0.00\% &0.00\%\\
		\hline
		
		{MF$\rightarrow$GB} &0.00\% &0.00\% &0.00\% &0.00\% &95.97\% &1.43\% &0.00\% &0.00\% &1.71\% &0.89\%\\
		\hline
		
		{GB$\rightarrow$MF} &0.00\% &0.00\% &0.00\% &0.00\% &0.78\% &93.67\% &0.92\% &3.56\% &1.07\% &0.00\%\\
		\hline
		
		{MF$\rightarrow$RS} &0.00\% &0.00\% &0.00\% &0.14\% &0.00\% &0.07\% &\underline{99.79\%} &0.00\% &0.00\% &0.00\%\\
		\hline
		
		{RS$\rightarrow$MF} & 0.00\% &1.32\% &0.00\% &0.00\% &0.00\% &10.46\% &1.74\% &85.22\% &1.26\% &0.00\%\\
		\hline
		
		{GB$\rightarrow$RS} & 0.00\% &0.00\% &0.00\% &0.00\% &0.28\% &0.14\% &0.28\% &2.20\% &96.88\% &0.21\%\\
		\hline
		
		{RS$\rightarrow$GB} &0.00\% &0.00\% &1.71\% &0.00\% &0.14\% &0.00\% &0.00\% &0.07\% &0.41\% &97.67\%\\
		\hline
		
		&\multicolumn{10}{c|}{\textbf{Two-Stream CNN\cite{liao2020robust}: ACC = 97.42\% {{TFLOPs = 0.82 Params = 20.23M}}}} \\
		\hline
		AU &\underline{99.93\%} &0.07\% &0.00\% &0.00\% &0.00\% &0.00\% &0.00\% &0.00\% &0.00\% &0.00\%\\ \hline
		MF &0.00\% &95.20\% &0.00\% &0.00\% &0.00\% &0.07\% &0.42\% &4.24\% &0.07\% &0.00\%\\ \hline
		GB &0.00\% &0.00\% &95.28\% &0.00\% &0.94\% &0.00\% &0.00\% &0.00\% &0.00\% &3.78\%\\ \hline
		RS &0.00\% &0.00\% &0.00\% &99.39\% &0.00\% &0.00\% &0.40\% &0.20\% &0.00\% &0.00\%\\ \hline
		MF$\rightarrow$GB &0.00\% &0.00\% &0.07\% &0.00\% &\textbf{98.77\%} &1.02\% &0.00\% &0.00\% &0.14\% &0.00\%\\ \hline
		GB$\rightarrow$MF &0.00\% &0.00\% &0.00\% &0.00\% &0.64\% &95.95\% &0.07\% &3.34\% &0.00\% &0.00\%\\ \hline
		MF$\rightarrow$RS &0.00\% &0.00\% &0.00\% &0.21\% &0.00\% &0.00\% &99.72\% &0.07\% &0.00\% &0.00\%\\ \hline
		RS$\rightarrow$MF &0.00\% &0.56\% &0.00\% &0.00\% &0.35\% &2.79\% &0.98\% &94.77\% &0.56\% &0.00\%\\ \hline
		GB$\rightarrow$RS &0.00\% &0.00\% &0.00\% &0.00\% &0.07\% &0.35\% &0.14\% &0.71\% &98.72\% &0.00\%\\ \hline
		RS$\rightarrow$GB &0.00\% &0.00\% &2.60\% &0.00\% &0.62\% &0.00\% &0.07\% &0.00\% &0.55\% &96.17\%\\
		\hline
		
		&\multicolumn{10}{c|}{\textbf{TransDetect\cite{you2022transformer}: ACC = 98.88\% {{TFLOPs = 1.69 Params = 86.04M}}}} \\
		\hline
		
		{AU} &\textbf{100.00\%} &0.00\% &0.00\% &0.00\% &0.00\% &0.00\% &0.00\% &0.00\% &0.00\% &0.00\%\\ \hline
		{MF} &0.00\% &\underline{98.76\%} &0.00\% &0.00\% &0.00\% &0.83\% &0.00\% &0.41\% &0.00\% &0.00\%\\ \hline
		{GB} &0.00\% &0.00\% &\textbf{97.83\%} &0.00\% &0.00\% &0.43\% &0.00\% &0.00\% &0.00\% &1.74\%\\ \hline
		{RS} &0.00\% &0.00\% &0.00\% &\textbf{100.00\%} &0.00\% &0.00\% &0.00\% &0.00\% &0.00\% &0.00\%\\ \hline
		{MF$\rightarrow$GB} &0.00\% &0.00\% &0.85\% &0.00\% &97.03\% &1.69\% &0.42\% &0.00\% &0.00\% &0.00\%\\ \hline
		{GB$\rightarrow$MF} &0.00\% &0.46\% &0.00\% &0.00\% &0.00\% &\underline{98.15\%} &0.00\% &1.39\% &0.00\% &0.00\%\\ \hline
		{MF$\rightarrow$RS} &0.00\% &0.00\% &0.00\% &0.00\% &0.00\% &0.00\% &\textbf{100.00\%} &0.00\% &0.00\% &0.00\%\\ \hline
		{RS$\rightarrow$MF} &0.00\% &0.00\% &0.00\% &0.00\% &0.00\% &0.80\% &0.40\% &\underline{98.80\%} &0.00\% &0.00\%\\ \hline
		{GB$\rightarrow$RS} &0.00\% &0.00\% &0.00\% &0.00\% &0.00\% &0.00\% &0.00\% &0.85\% &\underline{99.15\%} &0.00\%\\ \hline
		{RS$\rightarrow$GB} &0.00\% &0.00\% &1.21\% &0.00\% &0.00\% &0.00\% &0.00\% &0.00\% &0.00\% &\underline{98.79\%}\\ \hline
		&\multicolumn{10}{c|}{\textbf{Ours: ACC = 99.19\% {{TFLOPs = 1.08 Params = 42.60M}}}} \\
		\hline
		{AU}&99.78\%&0.00\%&0.06\%&0.00\%&0.09\%&0.00\%&0.00\%&0.06\%&0.00\%&0.00\% \\ \hline
		{MF}&0.00\%&\textbf{99.29\%}&0.00\%&0.00\%&0.00\%&0.70\%&0.00\%&0.00\%&0.00\%&0.00\% \\ \hline
		{GB}&0.00\%&0.00\%&\underline{97.70\%}&0.00\%&0.00\%&2.29\%&0.00\%&0.00\%&0.00\%&0.00\% \\ \hline
		{RS}&0.00\%&0.15\%&0.00\%&\underline{99.60\%}&0.00\%&0.00\%&0.00\%&0.00\%&0.24\%&0.00\% \\ \hline
		{MF$\rightarrow$GB}&0.00\%&0.00\%&1.37\%&0.00\%&\underline{98.25\%}&0.36\%&0.00\%&0.00\%&0.00\%&0.00\% \\ \hline
		{GB$\rightarrow$MF}&0.00\%&0.15\%&0.61\%&0.00\%&0.00\%&\textbf{99.23\%}&0.00\%&0.00\%&0.00\%&0.00\% \\ \hline
		{MF$\rightarrow$RS}&0.00\%&0.00\%&0.06\%&0.00\%&0.24\%&0.00\%&99.63\%&0.06\%&0.00\%&0.00\% \\ \hline
		{RS$\rightarrow$MF}&0.06\%&0.00\%&0.00\%&0.00\%&0.00\%&0.00\%&0.60\%&\textbf{99.32\%}&0.00\%&0.00\% \\ \hline
		{GB$\rightarrow$RS}&0.00\%&0.00\%&0.00\%&0.24\%&0.00\%&0.00\%&0.45\%&0.00\%&\textbf{99.29}\%&0.00\% \\ \hline
		{RS$\rightarrow$GB}&0.00\%&0.00\%&0.06\%&0.00\%&0.00\%&0.00\%&0.09\%&0.00\%&0.00\%&\textbf{99.84\%} \\ \hline
	\end{tabular}
	\label{tab:table_2}
	\vspace{-1em}
\end{table*}

We perform experiments on three public datasets, namely, UCID\cite{schaefer2003ucid}, BOSSbase\cite{bas2011break}, and RAISE\cite{dang2015raise}. The UCID dataset contains 1338 color images with sizes of $512\times384$ or $384\times512$, while both RAISE and BOSSbase consist of 10,000 high-resolution color images (approximately $3200\times 4800$).

\begin{table*}[t]
	\centering
	\caption{Confusion matrix of the operation chain detection using TMFNet on the RAISE dataset}
	\begin{tabular}{|c|c|c|c|c|c|c|c|c|c|c|}
		\hline
		\textbf{Operation}& {AU} & {MF} & {GB} & {RS} & {MF$\rightarrow$GB} & {GB$\rightarrow$MF} &{MF$\rightarrow$RS} & {RS$\rightarrow$MF} & {GB$\rightarrow$RS} & {RS$\rightarrow$GB} \\
		\hline
		&\multicolumn{10}{c|}{\textbf{TMFNet: ACC = 99.44\% {{TFLOPs = 1.08 Params = 42.60M}}}} \\
		
		\hline
		{AU} & 99.90\%&0.00\%	&0.00\%	&0.00\%	&0.06\%	&0.00\%	&0.00\%	&0.03\%	&0.00\%	&0.00\%\\
		\hline
		
		{MF} & 0.00\% & 99.23\% & 0.03\% & 0.00\% & 0.00\% &	0.06\% & 0.00\% & 0.09\% &	0.40\% & 0.09\% \\
		\hline
		
		{GB} & 0.00\% & 0.00\%	& 98.16\%	& 0.00\%	& 1.83\% & 0.00\% &0.00\%	&0.00\%	&0.00\%	&0.00\% \\
		\hline
		
		{RS} &0.00\% &0.00\% &0.00\% & 99.96\%	&0.00\%	&0.00\%	&0.00\%	&0.00\%	&0.03\%	&0.00\%\\
		\hline
		
		{MF$\rightarrow$GB} & 0.27\% & 0.00\%	&0.40\%	&0.00\%	&99.20\%	&0.12\%	&0.00\%	&0.00\%	&0.00\%	&0.00\%\\
		\hline
		
		{GB$\rightarrow$MF}& 0.00\% & 0.15\%	&0.43\%	&0.00\%	&0.00\%	&99.14\%	&0.27\%	&0.00\%	&0.00\%	&0.00\%\\
		\hline
		
		{MF$\rightarrow$RS} & 0.00\% & 0.00\%	&0.00\%	&0.00\%	&0.00\%	&0.00\%	&99.93\%	&0.06\%	&0.00\%	&0.00\%\\
		\hline
		
		{RS$\rightarrow$MF}& 0.00\% & 0.00\%	&0.00\%	&0.00\%	&0.00\%	&0.00\%	&0.00\%	&99.57\%	&0.43\%	&0.00\%\\
		\hline
		
		{GB$\rightarrow$RS}& 0.00\% & 0.24\%	&0.00\%	&0.15\%	&0.12\%	&0.00\%	&0.00\%	&0.00\%	&99.48\%	&0.00\%\\
		\hline
		
		{RS$\rightarrow$GB}& 0.00\% & 0.00\%	&0.00\%	&0.09\%	&0.00\%	&0.00\%	&0.00\%	&0.00\%	&0.00\%	&99.90\%\\
		\hline
	\end{tabular}
	\label{tab:table_10}
	\vspace{-1em}
\end{table*}

For the facilitate comparative experiments with previous methods, we segment the first 1,000 images in RAISE dataset into image patches of size $512 \times 512$ with a stride of 256. Therefore, we obtain approximately 32,000 image patches, in which 26,000 patches are for training, 3,000 patches are for validation, and the remaining patches are for testing. To evaluate the detection performance of our method on grayscale images, we transform those color image patches into grayscale ones. In the same way as color image patches, the grayscale patches are divided into training, validation, and testing sets. Then, we process the cropped data using the operations outlined in Eq. (2), which means that the size of train, valid, and test datasets after the operation is 10 times larger than the original dataset size. Additionally, we use the center cropping method to crop all images from the RAISE, BOSSBase, and UCID datasets to sizes of $256\times 256$, $128\times 128$, and $64\times 64$ patches for subsequent color image operation chain evaluation experiments, including robustness, generalization performance and transfer learning ability, etc.
\subsection{Operation Chain Detection of Non-JPEG Images}

\begin{figure}[t]
	\centering
		\begin{minipage}{0.48\linewidth}
		\centerline{\includegraphics[width=1.7in]{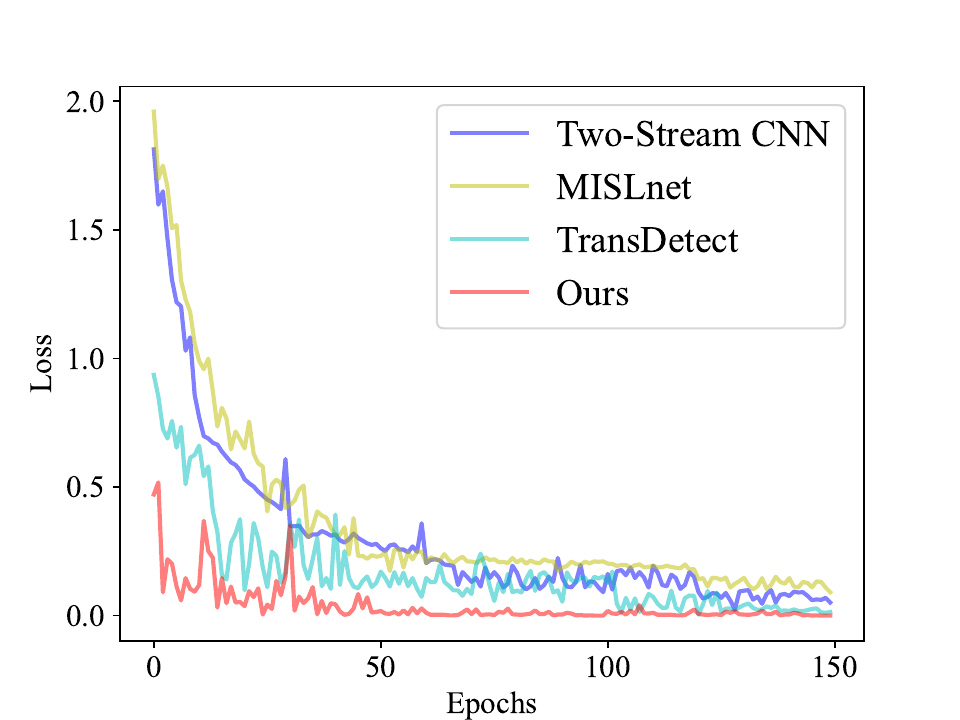}}
		\centerline{(a)}
	\end{minipage}
	\begin{minipage}{0.48\linewidth}
		\centerline{\includegraphics[width=1.7in]{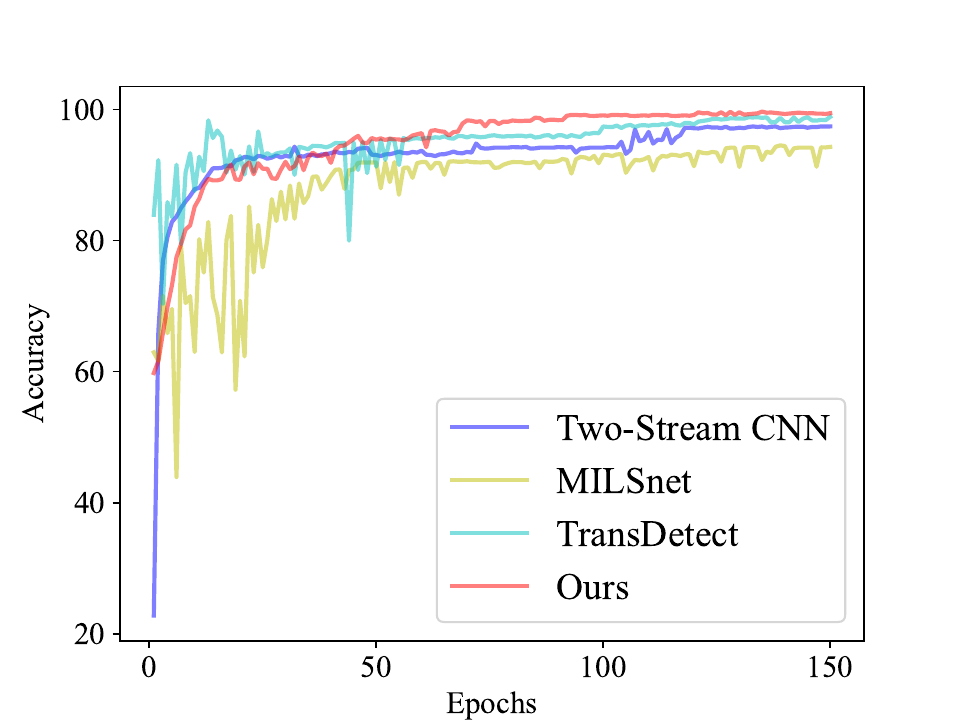}}
		\centerline{(b)}
	\end{minipage}
	\caption{The training curves of different methods on RAISE. (a) {validation losses} v.s. epochs. (b) ACC v.s. epochs.}
	\label{fig:subfig_10}
	\vspace{-1em}
\end{figure}

In this section, we evaluate the effectiveness of  method for operation chain detection under non JPEG compression condition. We assign a label to each operation chain and get 10 categories. The specific operation parameters constituting the operation chain are $O_{GB}^{1.1}$, $O_{MF}^5$, and $O_{RS}^{1.5}$, where subscript represents the operation type and superscript represents the specific parameters. For example, $O_{MF}^5$ denotes the median filtering operation with a filter kernel size of $5\times 5$. In the experiment, we select randomly and evenly from the data of each tag for training. We first conduct experiments on the $512\times 512$ grayscale image patches, and the confusion matrices for the operation chain detection using several state-of-the-art methods and the proposed TMFNet are shown in Table \ref{tab:table_2}. The first row and the first column represent the classification labels, the values on the diagonal represent the accuracy of correct classification of each category, and the non-diagonal region represents the error rate. From Table \ref{tab:table_2}, it can be seen that the detection accuracy of the method proposed in this study is better than that of MISLnet, Two-Stream CNN, and TransDetect in most categories on the grayscale image, with only a few categories having slightly lower detection accuracy than TransDetect. Overall, the average detection accuracy of MISLnet, Two-Stream CNN, and TransDetect is 94.40\%, 97.42\%, and 98.88\%, respectively, while the average detection accuracy of our method is 99.19\%.

To verify the effectiveness of our model in color image operation chain detection, we also present the detection accuracy of TMFNet on $512\times 512$ color image patches in Table \ref{tab:table_10}. The main difference between grayscale image operation chains and color image operation chains is that the former first converts to grayscale images and then performs operations, while the latter directly operates on color images. It is worth noting that by fusing the three-channel information, the operation chain average detection accuracy of TMFNet is superior to 99.44\%, which is significantly better than the previous three methods. {From these tables, in terms of computational complexity, our method takes only 1.08 TFLOPs, which is 0.61 TFLOPs less than the current state-of-the-art method TransDetect.} We also show the visualization results of the four models during the training process in Fig. \ref{fig:subfig_10}. Compared to the other three methods, our model exhibits a faster convergence rate and a smoother convergence process. {Since the official code of the TransDetect model is not publicly available, we reimplemented it according to the description in \cite{you2022transformer}.}

We also conduct experiments without any prior information to evaluate the robustness of the model. No prior information means that we know the operators involved in the operation chain, but the specific parameters of the operators are unknown. This situation is more realistic, because it is difficult to determine the specific parameters of the operators involved in the image operation in advance, but the parameter range of each operator can be estimated based on experience. We still use $O_{MF}$, $O_{GB}$, $O_{RS}$ and $O_{USM}$ as examples for experiments. We uniformly and randomly apply the operation parameters from Table \ref{tab:table_1} to process color images that have been cropped into $512\times 512$. Subsequently, we use the processed images for training, validation, and testing. The robustness test results are shown in Table \ref{tab:table_3}, and one can see that our method can effectively distinguish between different operation chains even without any prior information and achieve an average accuracy of 96.13\%, {which is 0.96\% higher than the second-best method, i.e., TransDetect.}

From the results in Tables \ref{tab:table_2}, \ref{tab:table_10} and \ref{tab:table_3}, it can be concluded that our method performs well in the operation chain detection on both grayscale and color images. This is due to the richer operation information in the color images we used, which amplifies the differences between different operation chains by combining global operation artifacts with residual noise information. At the same time, in the process of extracting residual noise from operations, we use RGB filters to aggregate the operation information of three channels, making our model more robust in the case of unknown priors.

\begin{table}[t]
	\centering
	\caption{Comparison performance without prior knowledge.}
	\renewcommand\arraystretch{1.5}
	\begin{tabular}{|c|c|c|c|c|}
		\hline
		\multirow{2}*{Operation}& \multicolumn{4}{c|}{Average ACC} \\
		\cline{2-5}
		&\cite{bayar2018constrained}& \cite{liao2020robust}	&{\cite{you2022transformer}}& Ours\\
		\hline
		
		{$O_{MF}, O_{GB}, O_{RS}$}& 88.07\% & 90.36\%	&{95.61\%}&\textbf{96.27\%}			\\
		\hline
		
		{$O_{MF}, O_{GB}, O_{USM}$}& 89.27\% & 90.11\%	&{95.94\%}&\textbf{97.33\%}		\\
		\hline
		
		{$O_{MF}, O_{USM}, O_{RS}$}& 86.91\% & 89.96\%	&{94.52\%}&\textbf{94.58\%}	\\
		\hline
		
		{$O_{USM}, O_{GB}, O_{RS}$}& 89.68\% & 90.62\%	&{94.70\%}&\textbf{96.36\%}		\\
		\hline
	\end{tabular}
	\label{tab:table_3}
	\vspace{-1.5em}
\end{table}

\begin{table}[h]
	\centering
	\setlength{\tabcolsep}{12pt}
	\caption{Comparison of robustness to JPEG compression.}
	\renewcommand\arraystretch{1.5}
	\begin{tabular}{|c|c|c|c|c|}
		\hline
		\multirow{2}*{QF}& \multicolumn{4}{c|}{Average ACC} \\
		\cline{2-5}
		& \cite{bayar2018constrained}& \cite{liao2020robust}&{\cite{you2022transformer}}& Ours\\
		\hline
		
		{70}& 81.07\% & 82.52\%	&{86.26\%}&\textbf{88.97\%}		\\
		\hline
		
		{75}& 82.27\% & 83.38\%	&{88.41\%}&\textbf{90.13\%}		\\
		\hline
		
		{80}& 85.32\% & 87.79\%	&{89.87\%}&\textbf{91.34\%}		\\
		\hline
		
		{85}& 89.56\% & 91.31\%	&{94.60\%}&\textbf{95.77\%}		\\
		\hline
		
		{90}& 91.68\% & 93.55\%	&{96.73\%}&\textbf{98.36\%}		\\
		\hline
	\end{tabular}
	\label{tab:table_4}
	\vspace{-1.5em}
\end{table}
\subsection{Operation Chain detection of JPEG Images}

\begin{table*}[htp]
	\centering
	\caption{Comparison of generalization on different block sizes on the RAISE dataset.}
	\renewcommand\arraystretch{1.5}
	\begin{tabular}{|c|c|c|c|c|c|c|c|c|c|}
		\hline
		\multirow{2}*{Methods} & \multicolumn{3}{c|}{{$O_{GB}^{1.1}$ AND {$O_{MF}^{5}$}}} & \multicolumn{3}{c|}{{$O_{GB}^{1.1}$ AND {$O_{RS}^{1.5}$}}} 	& \multicolumn{3}{c|}{{$O_{MF}$ AND {$O_{RS}$}}} \\	\cline{2-10}
		& $128\times 128$ & $256\times 256$ & $512\times 512$ & $128\times 128$ & $256\times 256$ & $512\times 512$ & {$128\times 128$} & {$256\times 256$} & {$512\times 512$} 	\\
		\hline
		{\cite{bayar2018constrained}}&87.62\%& 86.53\%& 85.12\%&86.35\%& 86.01\%& 84.79\%&{84.99\%}& {83.69\%}& {81.42\%}\\
		\hline
		{\cite{liao2020robust}}&89.06\%& 86.97\%& 86.29\%&90.09\%& 88.13\%& 87.28\%&{85.93\%}& {84.82\%}& {83.25\%}	\\
		\hline
		{\cite{you2022transformer}}&{86.93\%}& {86.19\%}& {85.06\%}&{88.82\%}& {86.20\%}& {86.11\%}&{84.68\%}& {84.13\%}& {83.29\%}	\\
		\hline
		{Ours}&\textbf{92.98\%}& \textbf{87.62\%}&\textbf{86.53\%}&\textbf{91.22\%}& \textbf{89.95\% }& \textbf{89.37\%}&\textbf{89.55\%}& \textbf{86.81\%}& \textbf{86.09\%} \\
		\hline
	\end{tabular}
	\label{tab:table_6}
	\vspace{-1em}
\end{table*}

In this section, we evaluate the effectiveness of the proposed method in the operation chain detection after JPEG compression. JPEG is a widely used image compression format, which compression method can minimize the number of storage bits, thus effectively reducing the file size. JPEG is a lossy compression where different quality factors (QF) can cause varying degrees of information loss. Essentially, JPEG compression can also be seen as an image operation, which can seriously interfere with the fingerprints left by previous operations and increase the difficulty of detection.

We subject the manipulated $256\times 256$ color image patches to JPEG compression and discuss the effect of different compression QFs on detection accuracy. We randomly and evenly send the JPEG compressed image to the trainer, and we test it after the training. From the experimental results in Table \ref{tab:table_4}, we can see that our proposed method can still accurately detect the operation chain compressed by JPEG. Note that, the detection accuracy decreases with the reduction of the compression QF value. This is because the smaller the QF value, the greater the information loss caused by compression, and the more serious the damage to the fingerprint left by previous operations. Compared with other methods, our method can still achieve 88.97\% accuracy even when QF=70, {which is 7.90\%, 6.45\% and 2.71\% higher than MISLnet, Two-Stream CNN and TransDetect,} respectively. Indicating that our model is more robust against JPEG compression.
\subsection{Evaluation of Model Generalization Ability}

In this section, we evaluate the generalization performance of the proposed method. Generalization performance refers to the test effect of the model on other datasets that are not involved in training. Because the imaging equipment of different data sets is different, the difference of imaging equipment will have an important impact on the constructed image data sets. When our model learns the operation characteristics of images, it is easy to fit these information into the model, resulting in the model relying on device information. Therefore, we often encounter the situation that we can get excellent performance on one dataset, but the detection accuracy on other datasets is far lower than expected.

\begin{table}[t]
	\centering
	\setlength{\tabcolsep}{3.5pt}
	\caption{Cross-dataset generalization performance evaluation.}
	\renewcommand\arraystretch{1.5}
	\begin{tabular}{|c|c|c|c|c|c|c|}
		\hline
		\multirow{2}*{Methods} & \multicolumn{2}{c|}{{$O_{GB}^{1.1}$ AND {$O_{MF}^{5}$}}}& \multicolumn{2}{c|}{{$O_{GB}^{1.1}$ AND {$O_{RS}^{1.5}$}}}& \multicolumn{2}{c|}{{$O_{MF}$ AND {$O_{RS}$}}} 	\\
		\cline{2-7}
		& {UCID} & {BossBase}& {UCID} & {BossBase}& {UCID} & {BossBase}	\\
		\hline
		{\cite{bayar2018constrained}}&90.09\%& 92.83\%&90.96\%& 93.14\%&{88.75\%}& {90.41\%}	\\
		\hline
		{\cite{liao2020robust}}&91.37\%& 92.85\%&92.88\%& 93.52\%&{90.06\%}& {90.92\%}	\\
		\hline
		{\cite{you2022transformer}}&{90.26\%}& {94.10\%}&{89.61\%}& {94.75\%}&{89.13\%}& {93.08\%}	\\
		\hline
		{Ours}&\textbf{94.24\%}& \textbf{96.33\%}&\textbf{94.08\%}& \textbf{95.62\%}&\textbf{92.86\%}& \textbf{94.55\%} \\
		\hline
		
\end{tabular}
\label{tab:table_5}
\vspace{-1em}
\end{table}

\begin{figure}[t]
	\centering
	\begin{minipage}{0.32\linewidth}
		\centerline{\includegraphics[width=1.16in]{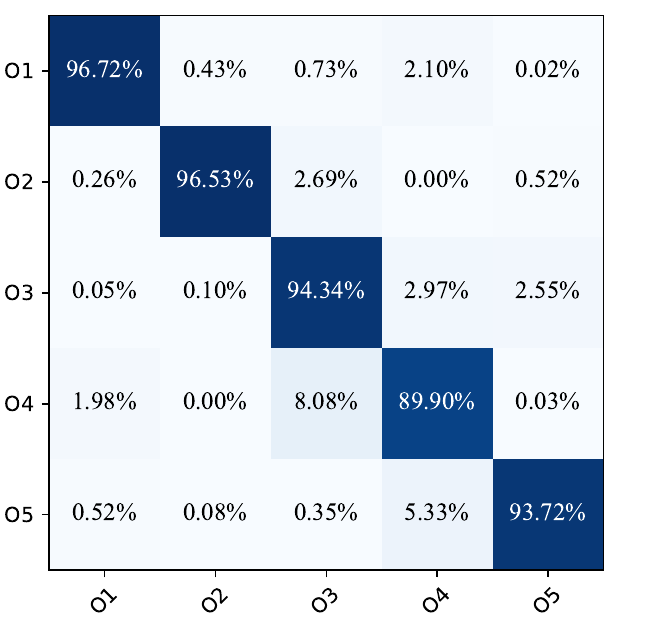}}
		\centerline{(a)}
	\end{minipage}
	\begin{minipage}{0.32\linewidth}
		\centerline{\includegraphics[width=1.18in]{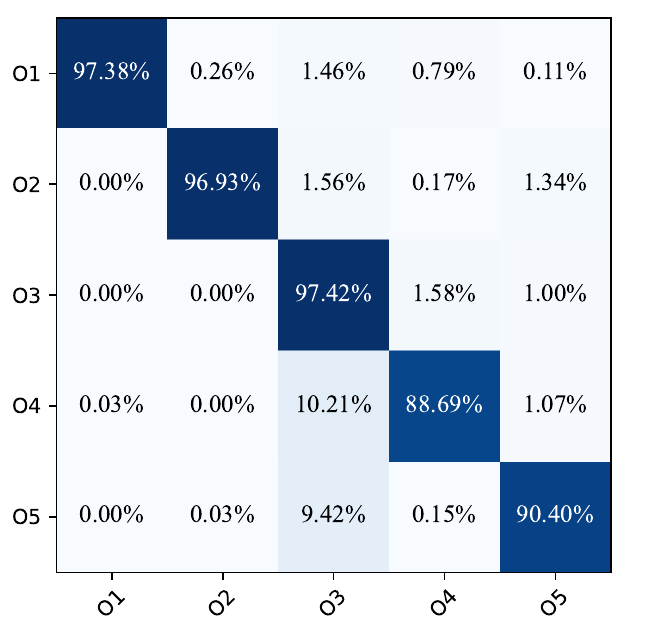}}
		\centerline{(b)}
	\end{minipage}
	\begin{minipage}{0.32\linewidth}
		\centerline{\includegraphics[width=1.15in]{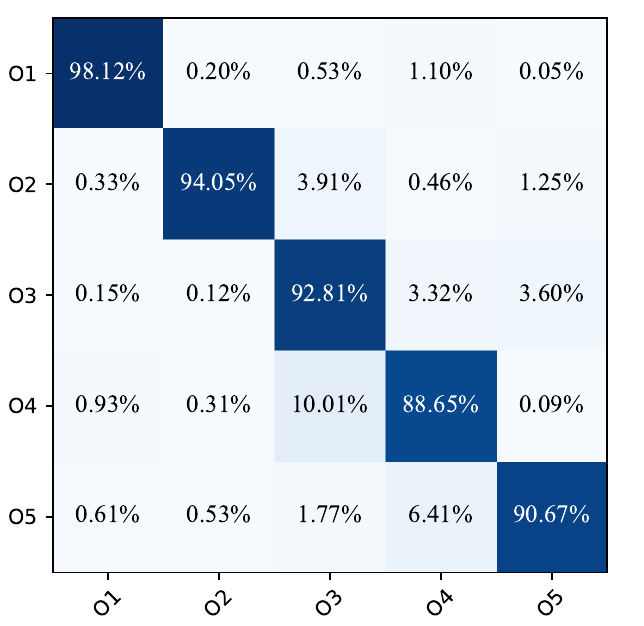}}
		\centerline{(c)}
	\end{minipage}
	\caption{Confusion matrices of generalization by using TMFNet on the UCID dataset. (a) $O_{GB}^{1.1}$ AND {$O_{MF}^{5}$}. (b) {$O_{GB}^{1.1}$ AND {$O_{RS}^{1.5}$}}. (c) {{$O_{MF}$ AND {$O_{RS}$}}}.}
	\label{fig:subfig_6}
	\vspace{-1.5em}
\end{figure}

We mainly evaluate the generalization ability of the model from two aspects of different data sets and different resolutions. We apply the parameters of the model trained on one dataset to other datasets for testing to evaluate the generalization ability of the model on different datasets. For example, we use the model trained on the RAISE dataset ($64\times 64$ color image patches) to test on UCID and bossbase datasets. We can also use a fixed resolution data set when training model parameters, and then use the trained model to test the data of other resolutions to evaluate the generalization ability of the model at different resolutions. For instance, you can train the model on the $64\times 64$ resolution RAISE training set, and then use the trained model to test on the $128\times 128$ and $256\times 256$ resolution RAISE test sets. For convenience, we summarize the average values of TMFNet generalization performance experiments in Tables \ref{tab:table_5} and \ref{tab:table_6}. Additionally, detailed experimental results of TMFNet on the UCID dataset are presented in the form of a confusion matrix in Fig. \ref{fig:subfig_6}, {where $O_n(n\in\{1,2,\dots,5\})$ represents the different operation chains}. One can observe that the generalization performance of our proposed method remains excellent. For example, in terms of cross-dataset generalization performance, {the overall average detection accuracy of our method on a chain consisting of different operations is improved by 3.58\%, 2.68\%, and 2.79\% compared to MISLnet, Two-Stream CNN, and TransDetect, respectively.}

\begin{table}
	\centering
	\caption{Comparison results of model transfer learning ability}
	\renewcommand\arraystretch{1.5}
	\begin{tabular}{|c|c|c|c|c|}
		\hline
		\multirow{2}*{Operation} & \multicolumn{4}{c|}{Average ACC} 	\\
		\cline{2-5}
		& {\cite{bayar2018constrained}} & {\cite{liao2020robust}}&{\cite{you2022transformer}} &{Ours} 	\\
		\hline
		{Source: {$O_{GB}^{1.1}$, {$O_{MF}^{5}$}}} &\multirow{2}*{93.20\%} & \multirow{2}*{95.35\%}& \multirow{2}*{97.74\%} &\multirow{2}*{ \textbf{98.60\%}}	\\
		Target: {$O_{GB}^{0.7}$, {$O_{MF}^{3}$}} & & && 	\\
		\hline
		{Source: {$O_{RS}^{1.2}$, {$O_{USM}$}}} &\multirow{2}*{93.59\%} & \multirow{2}*{94.24\%}& \multirow{2}*{97.31\%} &\multirow{2}*{ \textbf{98.13\%}}	\\
		Target: {$O_{RS}^{1.5}$, {$O_{USM}$}} & & & &	\\
		\hline
		{Source: {$O_{GB}^{1.1}$, {$O_{HE}$}}} &\multirow{2}*{94.63\%} & \multirow{2}*{95.98\%}& \multirow{2}*{98.15\%} &\multirow{2}*{ \textbf{99.05\%}}	\\
		Target: {$O_{GB}^{0.7}$, {$O_{HE}$}} & & & &	\\
		\hline
		{Source: {$O_{MF}^{5}$, {$O_{AWGN}$}}} &\multirow{2}*{94.09\%} & \multirow{2}*{96.30\%}& \multirow{2}*{96.92\%} &\multirow{2}*{ \textbf{98.91\%}}	\\
		Target: {$O_{MF}^{3}$, {$O_{AWGN}$}} & & & &	\\
		\hline
	\end{tabular}
	\label{tab:table_7}
	\vspace{-1.5em}
\end{table}

\subsection{Evaluation of Model Transfer Learning Ability}
In this section, we evaluate the transfer learning ability of the proposed method. Transfer learning is to study how to transfer knowledge between different but related tasks to help the learning of target tasks. Transfer learning makes the model avoid training from scratch and make full use of existing knowledge, so it can be better extended to new tasks. In the previous section, we discussed the robustness of the model. Different from directly using the trained model for testing, the performance evaluation of transfer learning needs fine-tuning training. Fine tuning refers to the adjustment based on the existing training model, rather than training a new model from scratch. During fine-tuning, we divide the initial learning rate by 10 and adjust the epoch to half of the original.

We only consider the transfer learning between the same operations. For example, we train the model on $O^{1.1}_{GB}$ and $O^5_{MF}$, and fine tune it on $O^{0.7}_{GB}$ and $O^{3}_{MF}$. In Table \ref{tab:table_7}, we show the results of the fine-tuning experiment on the RAISE dataset ($256\times 256$ color image patches). We will find that after fine-tuning, the accuracy of the experimental results can reach the level of training from scratch, and the time we spend is greatly reduced. This is very important to deal with the continuously diversified operation means and new tasks that may appear in the future. Our experimental results show that the fine-tuning model can still achieve good performance even in the face of similar tasks that have never been seen before.

As shown in Fig. \ref{fig:subfig_13} (a), by fine-tuning, we can make full use of the knowledge and feature representation ability we have learned in the pre-training model, so as to accelerate the convergence speed of the model and improve the performance. Fine-tuning enables the model to quickly adapt to new tasks and maintain good performance without a lot of training from scratch. This method of transfer learning is very effective for task transfer and new task processing in real applications. In the face of new operation means and tasks, our experimental results show that fine-tuning transfer learning can achieve excellent performance. This method can not only save time and resources, but also maintain the accuracy and robustness of the model. Therefore, the method proposed in our study have great potential in the field of transfer learning, can effectively deal with the challenges in transfer learning, and provide strong support for task transfer and processing in the future.
\vspace{-0.5em}

\subsection{Evaluation of Image Texture Suppression filters}
In this section, we evaluate the role of RGB filters and other methods in image operation chain detection and conduct experiments on non-JPEG compressed operation chain detection datasets ($256\times 256$ color image patches). Due to the use of preprocessing layers in MISLNet to suppress image textures, we use it as a comparison method in this section. We use SRM, CCL, and RGB filters as preprocessing layers for MISLNet and TMFNet, respectively, to evaluate the impact of these three texture suppression methods on the accuracy of image operation chain detection.

\begin{figure}[t]
	\centering
		\begin{minipage}{0.48\linewidth}
		\centerline{\includegraphics[width=1.7in]{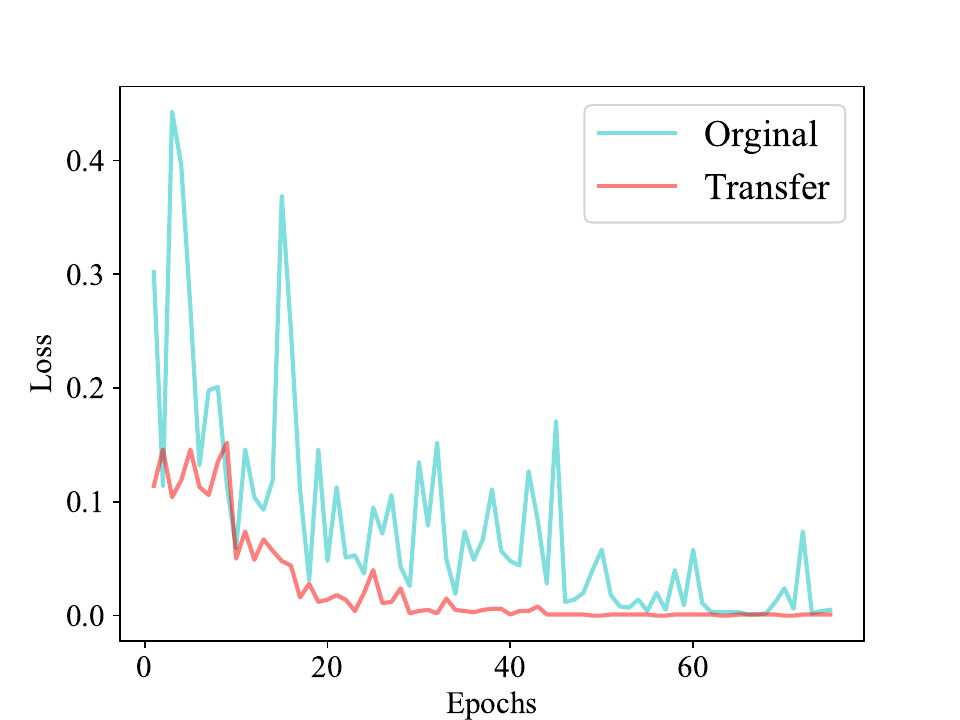}}
		\centerline{(a)}
	\end{minipage}
	\begin{minipage}{0.48\linewidth}
		\centerline{\includegraphics[width=1.7in]{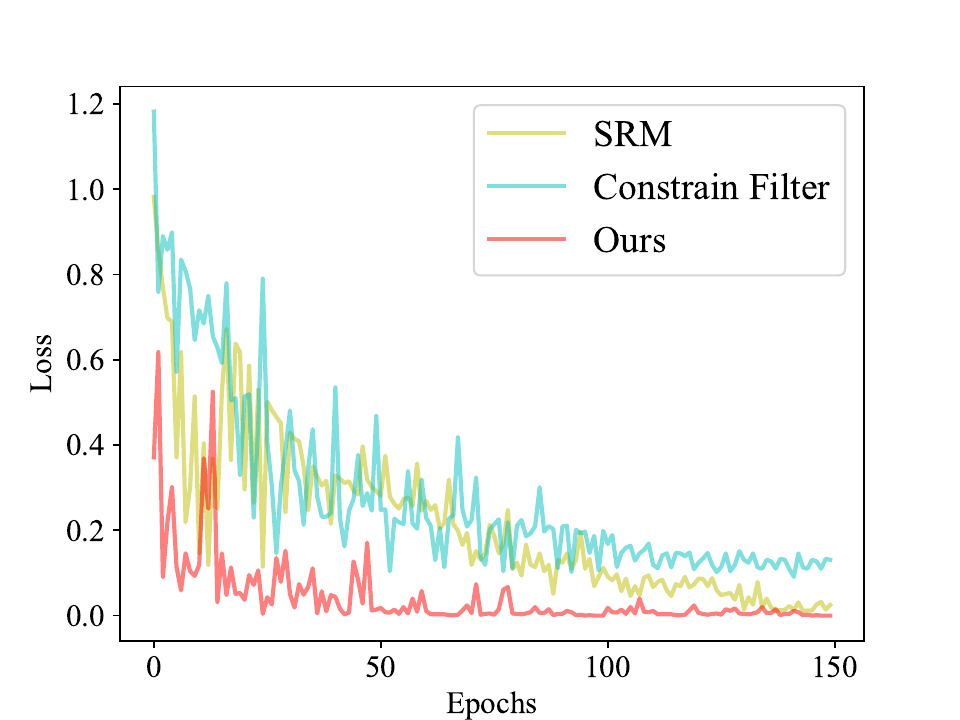}}
		\centerline{(b)}
	\end{minipage}
	\caption{Curves of training loss. (a) transfer learning. (b) different filters}
	\label{fig:subfig_13}
	\vspace{-0.5em}
\end{figure}
\begin{table}[t]
	\centering
	\caption{Comparison of three filters using MISLnet \cite{bayar2018constrained} and TMFNet on the RAISE dataset.}
	\renewcommand\arraystretch{1.5}
	\begin{tabular}{|c|c|c|c|}
		\hline
		\multirow{2}*{Methods} & \multicolumn{3}{c|}{Image Texture Suppression Method} 	\\
		\cline{2-4}
		& {CCL} & {SRM} & {RGB} 	\\
		\hline
		{\cite{bayar2018constrained}} &{94.30\%} & {93.16\%} &{ 96.73\%}	\\
		\hline
		{Ours} &{97.55\%} & {98.18\%} &{ \textbf{99.02\%}}	\\
		\hline
	\end{tabular}
	\label{tab:table_8}
	\vspace{-1.5em}
\end{table}
We present the results of the comparative experiment in Fig. \ref{fig:subfig_13} (b) and Table \ref{tab:table_8}. In Fig. \ref{fig:subfig_13} (b), we can observe that our proposed filter can make the convergence speed of the model faster than the other two filters. This means that our filter can guide the model to learn in the right direction more effectively in the process of suppressing texture. In contrast, the other two filters do not have the same advantages in terms of the convergence speed. In addition, in Table \ref{tab:table_8}, we compare the performance of the three filters in the test accuracy. The results show that our filter achieves higher accuracy than the other two filters on all test samples. Specifically, in our method, using RGB filters improved the average detection performance by 1.47\% and 0.85\%, respectively, compared to using SRM and CCL. This shows that our filter can suppress the image texture while maintaining the effective feature extraction ability for the target classification task.
\vspace{-0.5em}
\subsection{Ablation Study}
In order to evaluate the importance of each component in our proposed network, we conduct a series of ablation studies. We remove each component in the two-stream multi-channels fusion networks separately for experimental evaluation, and the comparison results are given in Table \ref{tab:table_9}. It can be seen that the noise residual stream with RGB filters shows superior performance compared to the noise residual stream without RGB filters and the independent spatial artifact stream, which proves that the RGB filter has a key role in improving the detection performance of the operation chain. In addition, the two-stream multi-channels fusion networks achieves the highest average classification accuracy, which indicates that the spatial artifact stream can provide a richer feature representation. Specifically, for $O_{GB}^{1.1}$ and $O_{MF}^{5}$ chains of image operators, the two-stream multi-channels fusion networks achieves an average classification accuracy improvement of 10.15\% and 8.37\%, respectively, compared to the network without RGB filters. Even when compared to the noisy residual stream with RGB filters, there is still a 2.61\% improvement in the average classification accuracy. Thus, spatial artifact streams and RGB filters play an important role in our proposed network. Moreover, we also study the impact of loss weight parameter settings on the detection performance of our model. Unlike\cite{you2022transformer},  which proposed a weighted cross-entropy loss function to address the tradeoff between category imbalance and length imbalance. In our study, the category of operation chain is balanced. In order to optimize the model efficiency, reduce parameter redundancy, and shorten the training period, we adopt a direct weighting strategy. One can see that the model with balanced weight coefficients exhibits a 0.39\% detection improvement compared to other settings (the highest accuracy among them is 98.36\%). Therefore, $\lambda_1 = \lambda_2 = 0.5$ are adopted.
\begin{table}[t]
	\centering
	\setlength{\tabcolsep}{3pt}
	\caption{Ablation Study of TMFNet.}
	\renewcommand\arraystretch{1.5}
	\begin{tabular}{|c|c|c|c|c|c|c|}
		\hline
		\multirow{2}*{Model}& \multirow{2}*{Spatial Stream} & \multicolumn{2}{c|}{Noise Stream}&\multicolumn{2}{c|}{Loss Weight}&\multirow{2}*{ACC} 	\\
		\cline{3-6}
		& & {RGB}& {Residual} &Other& Balance &\\
		\hline
		\multirow{5}*{TMFNet} &{\checkmark} &&&&&88.87\%\\
		\cline{2-7}
		& {} & &{\checkmark}&&&90.65\%\\
		\cline{2-7}
		& &{\checkmark}&{\checkmark}&&&96.41\%\\
		\cline{2-7}
		& {\checkmark}&{\checkmark}&{\checkmark}&{\checkmark}&&98.63\%\\
		\cline{2-7}
		& {\checkmark}&{\checkmark}&{\checkmark}&&{\checkmark}&\textbf{99.02\%}\\
		\hline
	\end{tabular}
	\label{tab:table_9}
	\vspace{-1.5em}
\end{table}
\begin{table}[t]
	\centering
	\setlength{\tabcolsep}{4pt}
	\caption{Ablation experiments using alternative network.}
	\renewcommand\arraystretch{1.5}
	{
		\begin{tabular}{|c|c|c|c|c|}
			\hline
			\multicolumn{2}{|c|}{Model}  & \multirow{2}*{ACC} & \multirow{2}*{TFLOPs}&\multirow{2}*{Params} \\
			\cline{1-2}
			Spatial& Noise &   &  &\\
			\hline
			Non-Pooling CNN& ShuffleNet\_v2\cite{ma2018shufflenet} & 96.37\% & 0.21&20.34M\\
			\hline
			Non-Pooling CNN& MobileNet\_v2\cite{sandler2018mobilenetv2} & 96.16\% & 0.22 &22.59M\\
			\hline
			Non-Pooling CNN& AlexNet\cite{krizhevsky2012imagenet} & 93.81\% & 0.24&80.22M \\
			\hline
			Non-Pooling CNN& VGG16\cite{simonyan2014very} & 97.52\% & 0.83&157.47M \\
			\hline
			Non-Pooling CNN& ResNet34 & 97.47\% & 0.36&40.36M \\
			\hline
			Non-Pooling CNN& ResNet101 & 99.07\% & 0.53 &61.59M\\
			\hline
			Pooling CNN& ResNet50 & 95.72\% & 0.30&35.18M \\
			\hline
			SRNet\cite{boroumand2018deep}& ResNet50 & 97.26\% & 0.36&88.64M \\
			\hline
			ViT\cite{dosovitskiy2020image}& ResNet50 & 98.89\% & 0.87&107.68M \\
			\hline
			Non-Pooling CNN& ResNet50 & 99.02\% & 0.38 &42.10M\\
			\hline
	\end{tabular}}
	\label{tab:table_11}
	\vspace{-0.5em}
\end{table}
\begin{figure*}[t]
	\centering
	\begin{minipage}{0.19\linewidth}
		\centerline{\includegraphics[width=1in]{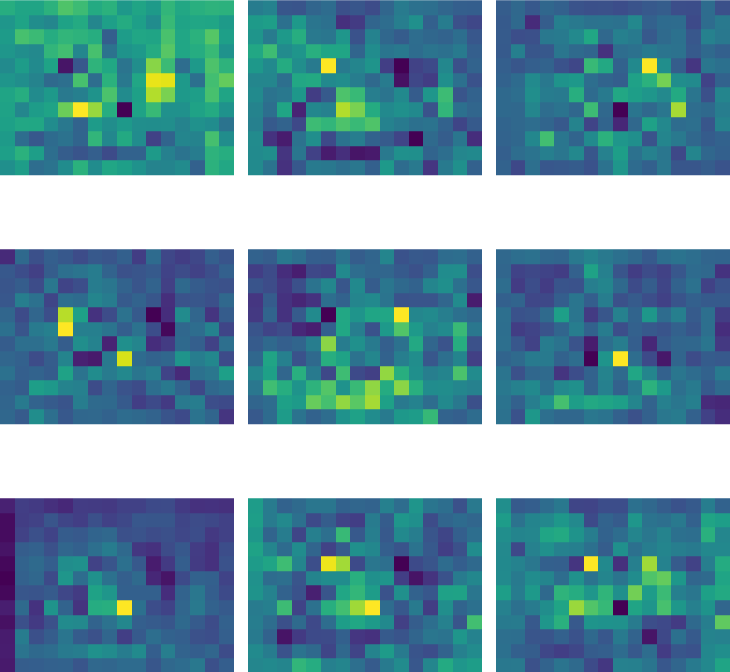}}
		\centerline{(a)}

	\end{minipage}
	\begin{minipage}{0.19\linewidth}
		\centerline{\includegraphics[width=1in]{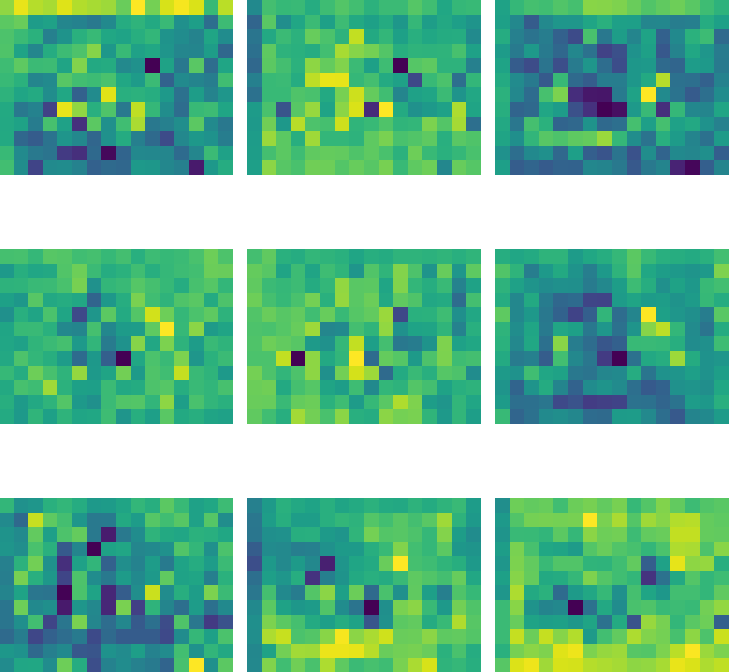}}
		\centerline{(b)}

	\end{minipage}
	\begin{minipage}{0.19\linewidth}
		\centerline{\includegraphics[width=1in]{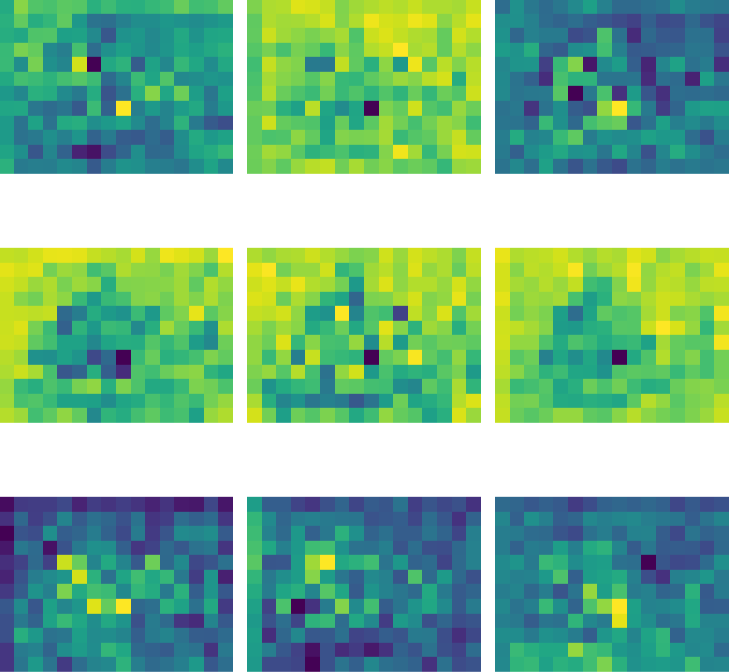}}
		\centerline{(c)}

	\end{minipage}
	\begin{minipage}{0.19\linewidth}
		\centerline{\includegraphics[width=1in]{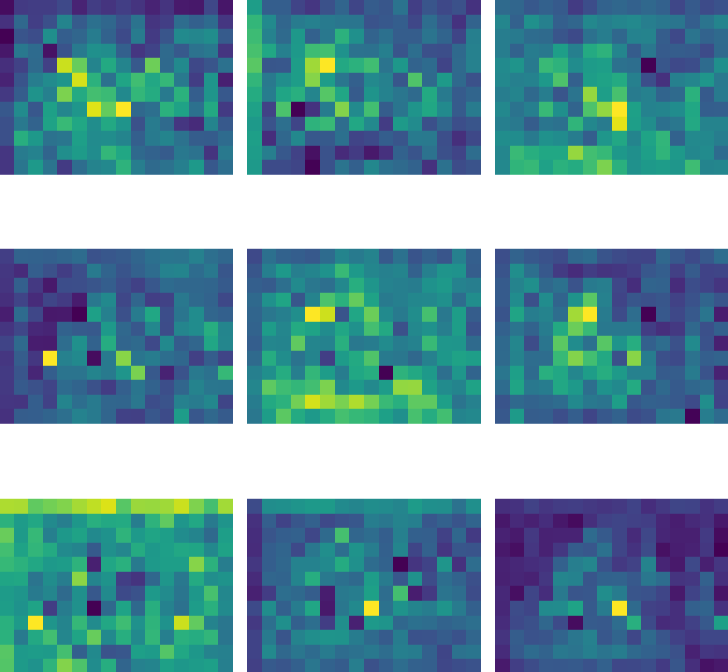}}
		\centerline{(d)}

	\end{minipage}
	\begin{minipage}{0.19\linewidth}
		\centerline{\includegraphics[width=1in]{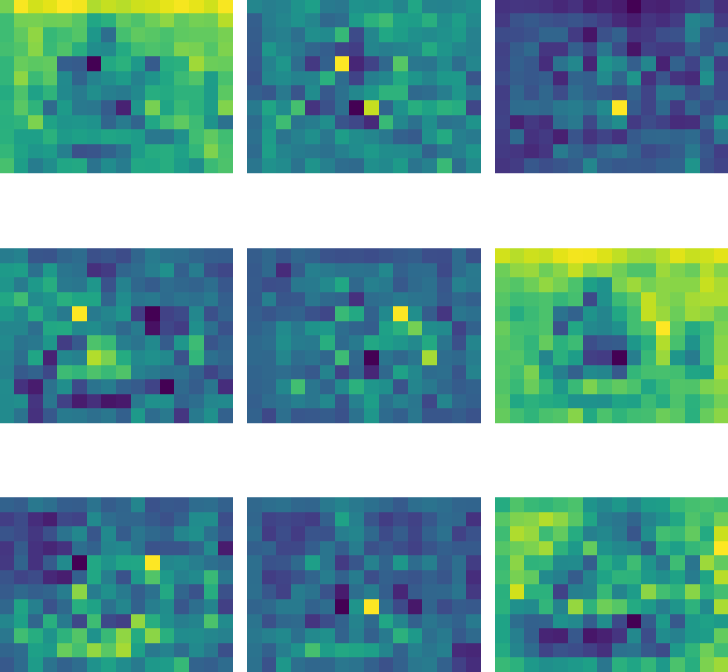}}
		\centerline{(e)}

	\end{minipage}\\
	\caption{Feature map visualization for different operation chains. (a) $O_{AU}$. (b) $O_{GB}^{1.0}$. (c) $O_{MF}^5$. (d) $O_{GB}^{1.0}$$\rightarrow$$O_{MF}^{5}$. (e) $O_{MF}^{5}$$\rightarrow$$O_{GB}^{1.0}$.}
	\label{fig:subfig_18}
	\vspace{-1em}
\end{figure*}
Based on the experimental results of the ablation study, we conclude that the network structure of noise residual stream without RGB filter layer can be regarded as a deeper spatial artifact stream, because they all adopt the idea of residual network construction. If we remove the RGB filters layer, the average classification accuracy obtained by the dual stream network will be close to the accuracy obtained by using only the spatial artifact stream. By fusing spatial artifact stream and noise residual stream, the global and local operation features of the operated image complement and reinforce each other, so that the network has stronger detection ability.

To analyze the characteristics of image operation chains, we performed feature visualization to show the characteristics of different image operation chains, as shown in Fig. \ref{fig:subfig_18}. We can see the energy distribution of the feature map for different operations is different to some extent. From Fig. \ref{fig:subfig_18} (a), for the original image, the energy is concentrated in the center of the feature maps. While, after a single operation,  the energy distribution is fairly uniform (see Fig. \ref{fig:subfig_18} (b) and (c)). In addition, for $O_{GB}^{1.0}$$\leftrightarrow$$O_{MF}^5$ chain, the energy distribution in several feature maps trends to these of the $O_{MF}^5$ operation. The reason is that the traces left by $O_{GB}^{1.0}$ could be severely weakened by later $O_{MF}^5$. As a result, the detection of long operation chains is a challenge for our proposed method.

\begin{table}[t]
	\centering
	\setlength{\tabcolsep}{4.5pt}
	\caption{Experiments on the known-length operation chain
		detection.}
	\renewcommand\arraystretch{1.5}
		\begin{tabular}{|c|c|c|c|c|c|c|}
			\hline 
			Methods & Metrics& $A_5^1$ & $A_5^2$ & $A_5^3$ & $A_5^4$ & $A_5^5$ \\
			\hline
			\multirow{3}{*}{\cite{you2022transformer}}& ACC & 100.00\%
			& 99.51\% & 97.01\% & 91.08\% & 87.81\%\\
			\cline{2-7}
			& ALMS & 100.00\% & 99.76\% & 98.60\% & 95.99\% & 94.11\%\\
			\cline{2-7}
			& BLEU & 1.0000 & 0.9979 & 0.9904 & 0.9645 & 0.9468\\
			\hline
			\multirow{3}{*}{TMFNet}& ACC & 100.00\% & 99.83\% & 96.89\% & 89.01\% &81.62\%\\
			\cline{2-7}
			& ALMS & 100.00\% & 99.91\% & 98.16\% & 93.62\% & 91.57\%\\
			\cline{2-7}
			& BLEU & 1.0000 & 0.9998 & 0.9839 & 0.9485 & 0.9329\\
			\hline
	\end{tabular}
	\label{tab:table_13}
		\vspace{-1.5em}
\end{table}
Furthermore, in order to verify the effectiveness of the two streams of the proposed model, we have conducted extensive ablation experiments using some existing mainstream networks (such as MobileNet\cite{sandler2018mobilenetv2}, ShuffleNet\cite{ma2018shufflenet}, AlexNET\cite{krizhevsky2012imagenet}, VGG166\cite{simonyan2014very}, SRNet\cite{boroumand2018deep}, ViT\cite{dosovitskiy2020image} and ResNet101 etc.) as the backbone of the two streams. Table \ref{tab:table_11} shows the detection accuracy, time complexity, and space complexity. We can see that Non-Pooling CNN+ResNet50 achieves better performance than all other combinations of the mainstream networks. Although Non-pooling CNN+ResNet101 obtains an improvement of 0.05\% on detection accuracy, both the time and space complexity increases by a factor of 1.5, which is a bit more than worth it. Similarly, in the spatial artifact stream, if the Non-Pooling CNN is replaced with the Vision Transform \cite{dosovitskiy2020image} (ViT), the complexity is doubled. Considering both the accuracy and complexity, we chose Non-pooling CNN + ResNet50 as the backbone network.
\vspace{-0.5em}

\subsection{Practical Applications}

In this section, we discuss the performance of the proposed model in practical applications, such as the detection of long chains and social media images. We first conduct experiments on long operation chains with two evaluation metrics, i.e., the longest matched subchain (ALMS)\cite{you2022transformer} and the Bilingual Evaluation Understudy (BLEU)\cite{papineni2002bleu}.  Tables \ref{tab:table_13} and \ref{tab:table_15} report the results of comparative detection with the current state-of-the-art detection method \cite{you2022transformer} under five operations (GB, MF, RS, JPEG, AWGN) for known lengths and unknown lengths (ranging from 1 to 5) respectively. In the scenario where the length of the operation chain is known, the proposed method achieves similar detection performance to the method \cite{you2022transformer} when the length is less than 3. However, as the length increases,  there is a certain degree of performance decrease. It indicates that, with the number of operation chains running into the hundreds, our classified-based approach is at a disadvantage. Under the condition of unknown length, the overall detect performance of our method is better than \cite{you2022transformer}, indicating better generalization. In a word, our method is able to detection short and long operation chains in both known and unknown lengths.


\begin{table}[t]
	\centering
	\caption{Comparison of two methods for detecting unknown-length operation chains.}
	\renewcommand\arraystretch{1.5}
		\begin{tabular}{|c|c|c|c|}
			\hline 
			Methods&  ACC & ALMS & BLEU   \\
			\hline
			\cite{you2022transformer}&79.06\% & 94.23\% & 0.9223   \\
			\hline
			TMFNet& 80.69\% & 91.57\%&0.9314\\
			\hline
	\end{tabular}
	\label{tab:table_15}\\
	\vspace{-1.5em}
\end{table}

{Furthermore, the deepfakes detection has become a hot topic in the field of image forensics. In practical application, the deepfake images are inevitably uploaded to social media platforms, where images may be performed by the operation chain. Therefore, we assess the effectiveness of the proposed method on deepfake images.

We randomly selected 10,000 images from the VGGFace2-HG dataset \cite{Cao18} for the operation chain evaluation experiments. The VGGFace2-HG dataset is a deepfake dataset containing more than one million $512\times512$ high-definition face-swapped images. These images are center cropped and adjusted their resolution to $256\times256$ and then 8,000 of them are used for training while the other 2,000 are for testing. As can be seen from Table \ref{tab:table_12}, our method achieves impressive results in terms of forensic accuracy. The average forensic accuracy of the five-category operation chain consisting of $O^{1.0}_{GB}$ and $O_{MF}^5$ is as high as 99.89\%. This is mainly due to the fact that each image in the VGGFace2-HG dataset contains rich face texture information. The RGB filter plays a great role in successfully aggregating the operation information of the three channels and suppressing the image texture. 
\begin{table}[t]
	\centering
	\setlength{\tabcolsep}{3.5pt}
	\caption{Experiments on the Deepfake dataset.}
	\renewcommand\arraystretch{1.5}
	{
		\begin{tabular}{|c|c|c|c|c|c|c|}
			\hline
			\multirow{2}{*}{Methods} & \multicolumn{5}{c|}{Operation}& ACC \\
			\cline{2-7}
			& AU & GB & MF & GB$\rightarrow$MF & MF$\rightarrow$GB &Mean\\
			\hline
			TMFNet& 100.00\% & 99.95\% & 99.95\% & 99.70\% & 99.85\% & 99.89\%\\
			\hline
	\end{tabular}}
	\label{tab:table_12}
	\vspace{-0.5em}
\end{table}

\begin{table}[t]
	\centering
	\setlength{\tabcolsep}{4pt}
	\caption{Comparison performance of sharing chain detection}
	\renewcommand\arraystretch{1.5}
		\begin{tabular}{|c|c|c|c|c|c|c|}
			\hline 
			\multirow{2}{*}{Methods} & \multicolumn{3}{c|}{R-SMUD} &\multicolumn{3}{c|}{V-SMUD}  \\
			\cline{2-7}
			&1&2&3&1&2&3\\
			\hline
			\cite{you2023image}&100.00\% &82.13\% &55.64\% &100.00\% &85.13\% &61.69\%\\
			\hline
			TMFNet&100.00\%&85.69\%&62.38\%&100.00\%&86.27\%&63.74\%\\
			\hline
	\end{tabular}
	\label{tab:table_16}
	\vspace{-1.5em}
\end{table}
	
We further explore the efficacy of the proposed method in detecting image sharing chains on social platforms. R-SMUD and V-SMUD \cite{phan2019tracking} are used as the experimental datasets where the images are shared no more than three times on 3 different platforms: Facebook (FB), Flickr (FL), and Twitter (TW). From the results in Table \ref{tab:table_16}, we can see that our method outperforms the recent method \cite{you2023image}, specially when the length is larger than 2. Indicating that our method has great potential for practical applications.


\section{Conclusion}
In this paper, we propose a novel deep learning framework consisting of a spatial artifact stream and a noise residual stream for color image operation chain detection. The spatial artifact stream can extract both shallow and deeper global operation artifacts. An RGB filter layer is first designed in the noise residual stream to suppress image texture and capture the low-level features. Then a multi-scale feature fusion is utilized for more comprehensively capturing high-level features. Finally, features from the two complementary streams are fed into a fusion module, which can effectively learn discriminative representations of operation chain artifacts. Extensive experiments have shown that the proposed method outperforms the state-of-the-art works in generalization, robustness, transfer learning ability, and texture suppression.


\nocite{*}
\bibliographystyle{unsrt}
\bibliography{reference}

\begin{thebibliography}{10}

\bibitem{popescu2005exposing}
Alin~C Popescu and Hany Farid.
\newblock Exposing digital forgeries by detecting traces of resampling.
\newblock {\em IEEE Transactions on signal processing}, 53(2):758--767, 2005.

\bibitem{zhang2020robustness}
Xianjin Liu, Wei Lu, Qin Zhang, Jiwu Huang, and Yun-Qing Shi.
\newblock Downscaling factor estimation on pre-{JPEG} compressed images.
\newblock {\em IEEE Transactions on Circuits and Systems for Video Technology},
  30(3):618--631, 2019.

\bibitem{vazquez2017random}
David V{\'a}zquez-Pad{\'\i}n, Fernando P{\'e}rez-Gonz{\'a}lez, and Pedro
  Comesana-Alfaro.
\newblock A random matrix approach to the forensic analysis of upscaled images.
\newblock {\em IEEE Transactions on Information Forensics and Security},
  12(9):2115--2130, 2017.

\bibitem{chen2015median}
Jiansheng Chen, Xiangui Kang, Ye~Liu, and Z~Jane Wang.
\newblock Median filtering forensics based on convolutional neural networks.
\newblock {\em IEEE Signal Processing Letters}, 22(11):1849--1853, 2015.

\bibitem{tang2018median}
Hongshen Tang, Rongrong Ni, Yao Zhao, and Xiaolong Li.
\newblock Median filtering detection of small-size image based on {CNN}.
\newblock {\em Journal of Visual Communication and Image Representation},
  51:162--168, 2018.

\bibitem{yuan2011blind}
Hai-Dong Yuan.
\newblock Blind forensics of median filtering in digital images.
\newblock {\em IEEE Transactions on Information Forensics and Security},
  6(4):1335--1345, 2011.

\bibitem{chen2013blind}
Chenglong Chen, Jiangqun Ni, and Jiwu Huang.
\newblock Blind detection of median filtering in digital images: A difference
  domain based approach.
\newblock {\em IEEE Transactions on Image Processing}, 22(12):4699--4710, 2013.

\bibitem{stamm2010forensic}
Matthew~C Stamm and KJ~Ray Liu.
\newblock Forensic detection of image manipulation using statistical intrinsic
  fingerprints.
\newblock {\em IEEE Transactions on Information Forensics and Security},
  5(3):492--506, 2010.

\bibitem{cao2014contrast}
Gang Cao, Yao Zhao, Rongrong Ni, and Xuelong Li.
\newblock Contrast enhancement-based forensics in digital images.
\newblock {\em IEEE transactions on information forensics and security},
  9(3):515--525, 2014.

\bibitem{de2015second}
Xiuli Bi, Yixuan Shang, Bo~Liu, Bin Xiao, Weisheng Li, and Xinbo Gao.
\newblock A versatile detection method for various contrast enhancement
  manipulations.
\newblock {\em IEEE Transactions on Circuits and Systems for Video Technology},
  33(2):491--504, 2022.

\bibitem{cao2011unsharp}
Gang Cao, Yao Zhao, Rongrong Ni, and Alex~C Kot.
\newblock Unsharp masking sharpening detection via overshoot artifacts
  analysis.
\newblock {\em IEEE Signal Processing Letters}, 18(10):603--606, 2011.

\bibitem{ding2014novel}
Feng Ding, Guopu Zhu, and Yun~Qing Shi.
\newblock A novel method for detecting image sharpening based on local binary
  pattern.
\newblock In {\em {Digital-Forensics and Watermarking: 12th International
  Workshop}}, pages 180--191. Springer, 2014.

\bibitem{ding2014edge}
Feng Ding, Guopu Zhu, Jianquan Yang, Jin Xie, and Yun-Qing Shi.
\newblock Edge perpendicular binary coding for {USM} sharpening detection.
\newblock {\em IEEE signal processing letters}, 22(3):327--331, 2014.

\bibitem{fan2003identification}
Zhigang Fan and Ricardo~L De~Queiroz.
\newblock Identification of bitmap compression history: {JPEG} detection and
  quantizer estimation.
\newblock {\em IEEE Transactions on Image Processing}, 12(2):230--235, 2003.

\bibitem{luo2010jpeg}
Yakun Niu, Xiaolong Li, Yao Zhao, and Rongrong Ni.
\newblock Detection of double {JPEG} compression with the same quantization
  matrix via convergence analysis.
\newblock {\em IEEE Transactions on Circuits and Systems for Video Technology},
  32(5):3279--3290, 2021.

\bibitem{wang2021detecting}
Hao Wang, Jinwei Wang, Xiangyang Luo, Yuhui Zheng, Bin Ma, Jinsheng Sun, and
  Sunil~Kr Jha.
\newblock Detecting aligned double {JPEG} compressed color image with same
  quantization matrix based on the stability of image.
\newblock {\em IEEE Transactions on Circuits and Systems for Video Technology},
  32(6):4065--4080, 2021.

\bibitem{qiu2014universal}
Xiaoqing Qiu, Haodong Li, Weiqi Luo, and Jiwu Huang.
\newblock A universal image forensic strategy based on steganalytic model.
\newblock In {\em {Proceedings of the 2nd ACM Workshop on Information Hiding
  and Multimedia Security}}, pages 165--170, 2014.

\bibitem{fridrich2012rich}
Jessica Fridrich and Jan Kodovsky.
\newblock Rich models for steganalysis of digital images.
\newblock {\em IEEE Transactions on information Forensics and Security},
  7(3):868--882, 2012.

\bibitem{li2016identification}
Haodong Li, Weiqi Luo, Xiaoqing Qiu, and Jiwu Huang.
\newblock Identification of various image operations using residual-based
  features.
\newblock {\em IEEE Transactions on Circuits and Systems for Video Technology},
  28(1):31--45, 2016.

\bibitem{li2022image}
Xiaohong Liu, Yaojie Liu, Jun Chen, and Xiaoming Liu.
\newblock {PSCC}-{N}et: Progressive spatio-channel correlation network for
  image manipulation detection and localization.
\newblock {\em IEEE Transactions on Circuits and Systems for Video Technology},
  32(11):7505--7517, 2022.

\bibitem{cozzolino2015efficient}
Yulan Zhang, Guopu Zhu, Xing Wang, Xiangyang Luo, Yicong Zhou, Hongli Zhang,
  and Ligang Wu.
\newblock {CNN-Transformer Based Generative Adversarial Network for Copy-Move
  Source/Target Distinguishment}.
\newblock {\em IEEE Transactions on Circuits and Systems for Video Technology},
  33(5):2019--2032, 2022.

\bibitem{rao2020deep}
Xudong Zhao, Shilin Wang, Shenghong Li, and Jianhua Li.
\newblock Passive image-splicing detection by a 2-{D} noncausal markov model.
\newblock {\em IEEE Transactions on Circuits and Systems for Video Technology},
  25(2):185--199, 2014.

\bibitem{krishnamoorthy2022splicing}
Yulan Zhang, Guopu Zhu, Ligang Wu, Sam Kwong, Hongli Zhang, and Yicong Zhou.
\newblock Multi-task {SE}-network for image splicing localization.
\newblock {\em IEEE Transactions on Circuits and Systems for Video Technology},
  32(7):4828--4840, 2021.

\bibitem{amerini2017localization}
Irene Amerini, Tiberio Uricchio, Lamberto Ballan, and Roberto Caldelli.
\newblock Localization of {JPEG} double compression through multi-domain
  convolutional neural networks.
\newblock In {\em {2017 IEEE Conference on Computer Vision and Pattern
  Recognition Workshops (CVPRW)}}, pages 1865--1871. IEEE, 2017.

\bibitem{bayar2018constrained}
Belhassen Bayar and Matthew~C Stamm.
\newblock Constrained convolutional neural networks: {A} new approach towards
  general purpose image manipulation detection.
\newblock {\em IEEE Transactions on Information Forensics and Security},
  13(11):2691--2706, 2018.

\bibitem{barni2018cnn}
Mauro Barni, Andrea Costanzo, Ehsan Nowroozi, and Benedetta Tondi.
\newblock {CNN}-based detection of generic contrast adjustment with {JPEG}
  post-processing.
\newblock In {\em {2018 25th IEEE International Conference on Image Processing
  (ICIP)}}, pages 3803--3807. IEEE, 2018.

\bibitem{chen2023identification}
Jiaxin Chen, Xin Liao, Wei Wang, and Zheng Qin.
\newblock Identification of image global processing operator chain based on
  feature decoupling.
\newblock {\em Information Sciences}, 637:118961, 2023.

\bibitem{you2021transformer}
Jiaxiang You, Yuanman Li, Jiantao Zhou, Zhongyun Hua, Weiwei Sun, and Xia Li.
\newblock A transformer based approach for image manipulation chain detection.
\newblock In {\em {Proceedings of the 29th ACM International Conference on
  Multimedia}}, pages 3510--3517, 2021.

\bibitem{you2022transformer}
Yuanman Li, Jiaxiang You, Jiantao Zhou, Wei Wang, Xin Liao, and Xia Li.
\newblock {Image Operation Chain Detection with Machine Translation Framework}.
\newblock {\em IEEE Transactions on Multimedia}, 25:6852--6867, 2023.

\bibitem{liao2020robust}
Xin Liao, Kaide Li, Xinshan Zhu, and KJ~Ray Liu.
\newblock Robust detection of image operator chain with two-stream
  convolutional neural network.
\newblock {\em IEEE Journal of Selected Topics in Signal Processing},
  14(5):955--968, 2020.

\bibitem{verde2023multi}
Sebastiano Verde, Cecilia Pasquini, Federica Lago, Alessandro Goller, Francesco
  De~Natale, Alessandro Piva, and Giulia Boato.
\newblock Multi-clue reconstruction of sharing chains for social media images.
\newblock {\em IEEE Transactions on Multimedia}, 25:9491--9505, 2023.

\bibitem{liao2020adaptive}
Xin Liao, Jiaojiao Yin, Mingliang Chen, and Zheng Qin.
\newblock Adaptive payload distribution in multiple images steganography based
  on image texture features.
\newblock {\em IEEE Transactions on Dependable and Secure Computing},
  19(2):897--911, 2020.

\bibitem{guo2023exposing}
Zhiqing Guo, Gaobo Yang, Jiyou Chen, and Xingming Sun.
\newblock Exposing deepfake face forgeries with guided residuals.
\newblock {\em IEEE Transactions on Multimedia}, 25:8458--8470, 2023.

\bibitem{zhou2018learning}
Peng Zhou, Xintong Han, Vlad~I Morariu, and Larry~S Davis.
\newblock Learning rich features for image manipulation detection.
\newblock In {\em {Proceedings of the IEEE Conference on Computer Vision and
  Pattern Recognition}}, pages 1053--1061, 2018.

\bibitem{dang2015raise}
Duc-Tien Dang-Nguyen, Cecilia Pasquini, Valentina Conotter, and Giulia Boato.
\newblock "{RAISE}: A raw images dataset for digital image forensics".
\newblock In {\em {Proceedings of the 6th ACM Multimedia Systems Conference}},
  pages 219--224, 2015.

\bibitem{boroumand2018deep}
Mehdi Boroumand, Mo~Chen, and Jessica Fridrich.
\newblock Deep residual network for steganalysis of digital images.
\newblock {\em IEEE Transactions on Information Forensics and Security},
  14(5):1181--1193, 2018.

\bibitem{he2016deep}
Kaiming He, Xiangyu Zhang, Shaoqing Ren, and Jian Sun.
\newblock Deep residual learning for image recognition.
\newblock In {\em {Proceedings of the IEEE Conference on Computer Vision and
  Pattern Recognition}}, pages 770--778, 2016.

\bibitem{schaefer2003ucid}
Gerald Schaefer and Michal Stich.
\newblock {UCID}: An uncompressed color image database.
\newblock In {\em Storage and retrieval methods and applications for multimedia
  2004}, volume 5307, pages 472--480. SPIE, 2003.

\bibitem{bas2011break}
Patrick Bas, Tom{\'a}{\v{s}} Filler, and Tom{\'a}{\v{s}} Pevn{\`y}.
\newblock 'break our steganographic system': the ins and outs of organizing
  {BOSS}.
\newblock In {\em {International Workshop on Information Hiding}}, pages
  59--70. Springer, 2011.

\bibitem{sandler2018mobilenetv2}
Mark Sandler, Andrew Howard, Menglong Zhu, Andrey Zhmoginov, and Liang-Chieh
  Chen.
\newblock Mobilenetv2: Inverted residuals and linear bottlenecks.
\newblock In {\em {Proceedings of the IEEE Conference on Computer Vision and
  Pattern Recognition}}, pages 4510--4520, 2018.

\bibitem{ma2018shufflenet}
Ningning Ma, Xiangyu Zhang, Hai-Tao Zheng, and Jian Sun.
\newblock Shufflenet v2: Practical guidelines for efficient cnn architecture
  design.
\newblock In {\em {Proceedings of the European Conference on Computer Vision
  (ECCV)}}, pages 116--131, 2018.

\bibitem{krizhevsky2012imagenet}
Alex Krizhevsky, Ilya Sutskever, and Geoffrey~E Hinton.
\newblock Imagenet classification with deep convolutional neural networks.
\newblock {\em {Advances in Neural Information Processing Systems}},
  25:1097--1105, 2012.

\bibitem{simonyan2014very}
Karen Simonyan and Andrew Zisserman.
\newblock Very deep convolutional networks for large-scale image recognition.
\newblock {\em arXiv preprint arXiv:1409.1556}, 2014.

\bibitem{dosovitskiy2020image}
Alexey Dosovitskiy, Lucas Beyer, Alexander Kolesnikov, Dirk Weissenborn,
  Xiaohua Zhai, Thomas Unterthiner, Mostafa Dehghani, Matthias Minderer, Georg
  Heigold, Sylvain Gelly, et~al.
\newblock An image is worth 16x16 words: Transformers for image recognition at
  scale.
\newblock {\em arXiv preprint arXiv:2010.11929}, 2020.

\bibitem{papineni2002bleu}
Kishore Papineni, Salim Roukos, Todd Ward, and Wei-Jing Zhu.
\newblock {BLEU}: a method for automatic evaluation of machine translation.
\newblock In {\em {Proceedings of the 40th annual meeting of the Association
  for Computational Linguistics}}, pages 311--318, 2002.

\bibitem{Cao18}
Liu Naiyuan.
\newblock {VGGFace2-HQ}: A dataset for recognising faces across pose and age -
  high quality.
\newblock In {\em https://github.com/NNNNAI/VGGFace2-HQ}, 2023.

\bibitem{phan2019tracking}
Quoc-Tin Phan, Giulia Boato, Roberto Caldelli, and Irene Amerini.
\newblock Tracking multiple image sharing on social networks.
\newblock In {\em {IEEE International Conference on Acoustics, Speech and
  Signal Processing (ICASSP)}}, pages 8266--8270. IEEE, 2019.

\bibitem{you2023image}
Jiaxiang You, Yuanman Li, Rongqin Liang, Yuxuan Tan, Jiantao Zhou, and Xia Li.
\newblock Image sharing chain detection via sequence-to-sequence model.
\newblock In {\em {IEEE International Conference on Acoustics, Speech and
  Signal Processing (ICASSP)}}, pages 1--5. IEEE, 2023.

\end{thebibliography}

\begin{IEEEbiography}[{\includegraphics[width=1in,height=1.25in,clip,keepaspectratio]{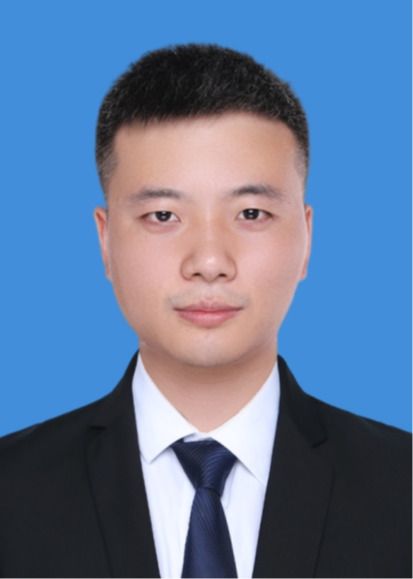}}]{Yakun Niu} received the Ph.D. degree from the Institute of Information Science, Beijing Jiaotong University, Beijing, China, in 2021. He is currently a Lecturer with the School of Computer and Information Engineering, Henan University, Kaifeng, China. His current research interests include multimedia security and forensics.\end{IEEEbiography}

\begin{IEEEbiography}[{\includegraphics[width=1in,height=1.25in,clip,keepaspectratio]{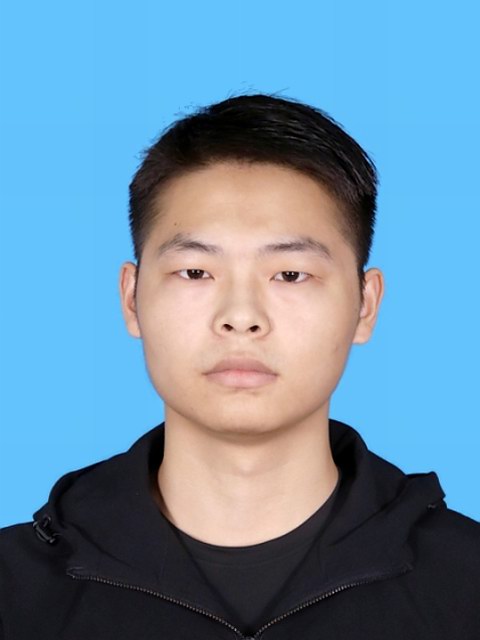}}]{Lei Tan} received a Bachelor's degree in Digital Media Technology from Henan University of Animal Husbandry and Economy in 2021 and is currently a Master's student at the School of Computer and Information Engineering, Henan University. His current research interests include operation forensics in image forensics.\end{IEEEbiography}

\begin{IEEEbiography}[{\includegraphics[width=1in,height=1.25in,clip,keepaspectratio]{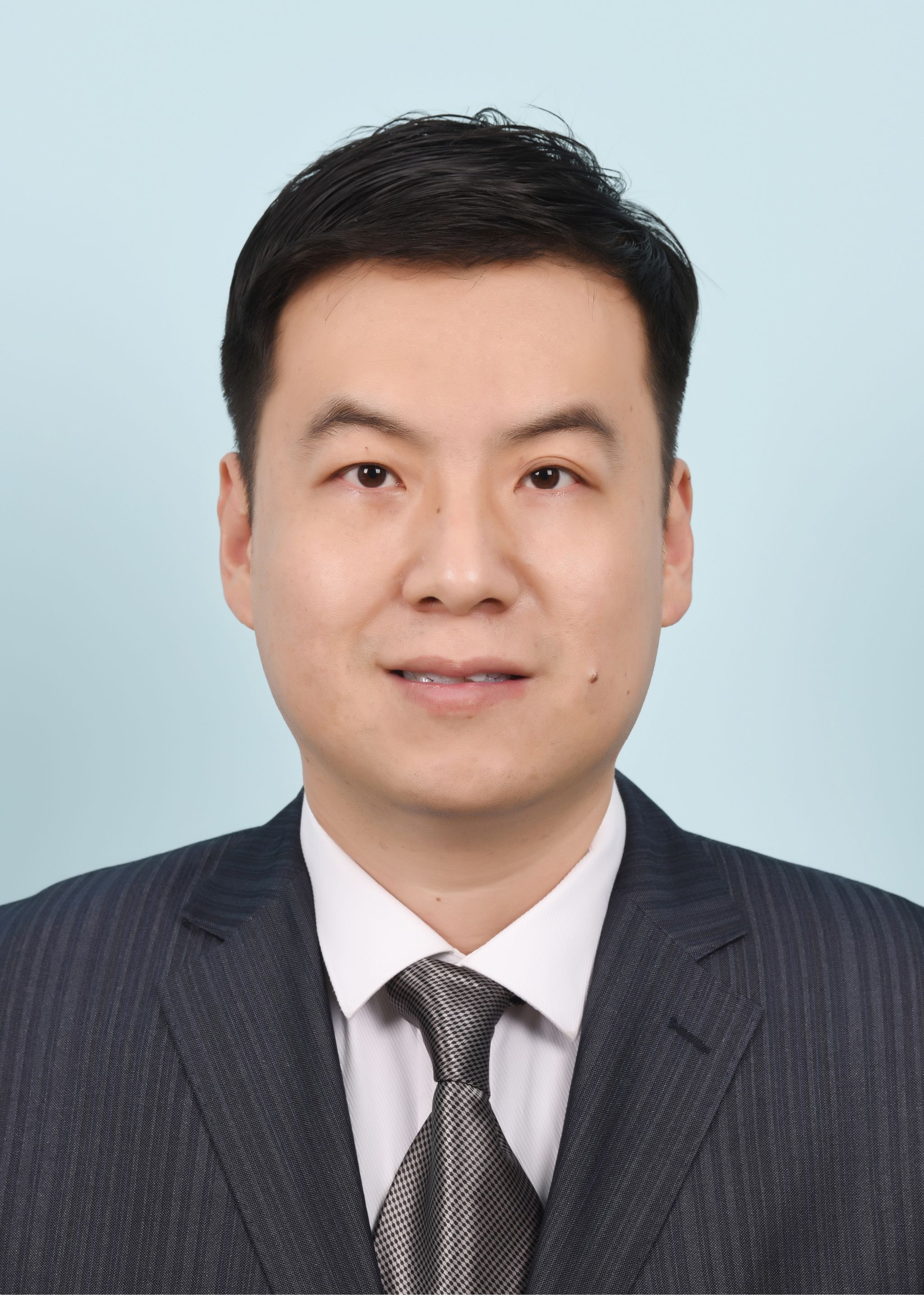}}]{Lei Zhang}received the B.S. and M.S. degrees from Henan University Kaifeng China in 2003 and 2006, respectively, and the Ph.D. degree in computer science and technology from the Harbin Institute of Technology, Harbin, China, in 2015. He is an Associate Professor with the Institute of Data and Knowledge Engineering, School of Computer and Information Engineering, Henan University, China. His research interests include deep learning, computer networks, and cyberspace security.\end{IEEEbiography}

\begin{IEEEbiography}[{\includegraphics[width=1in,height=1.25in,clip,keepaspectratio]{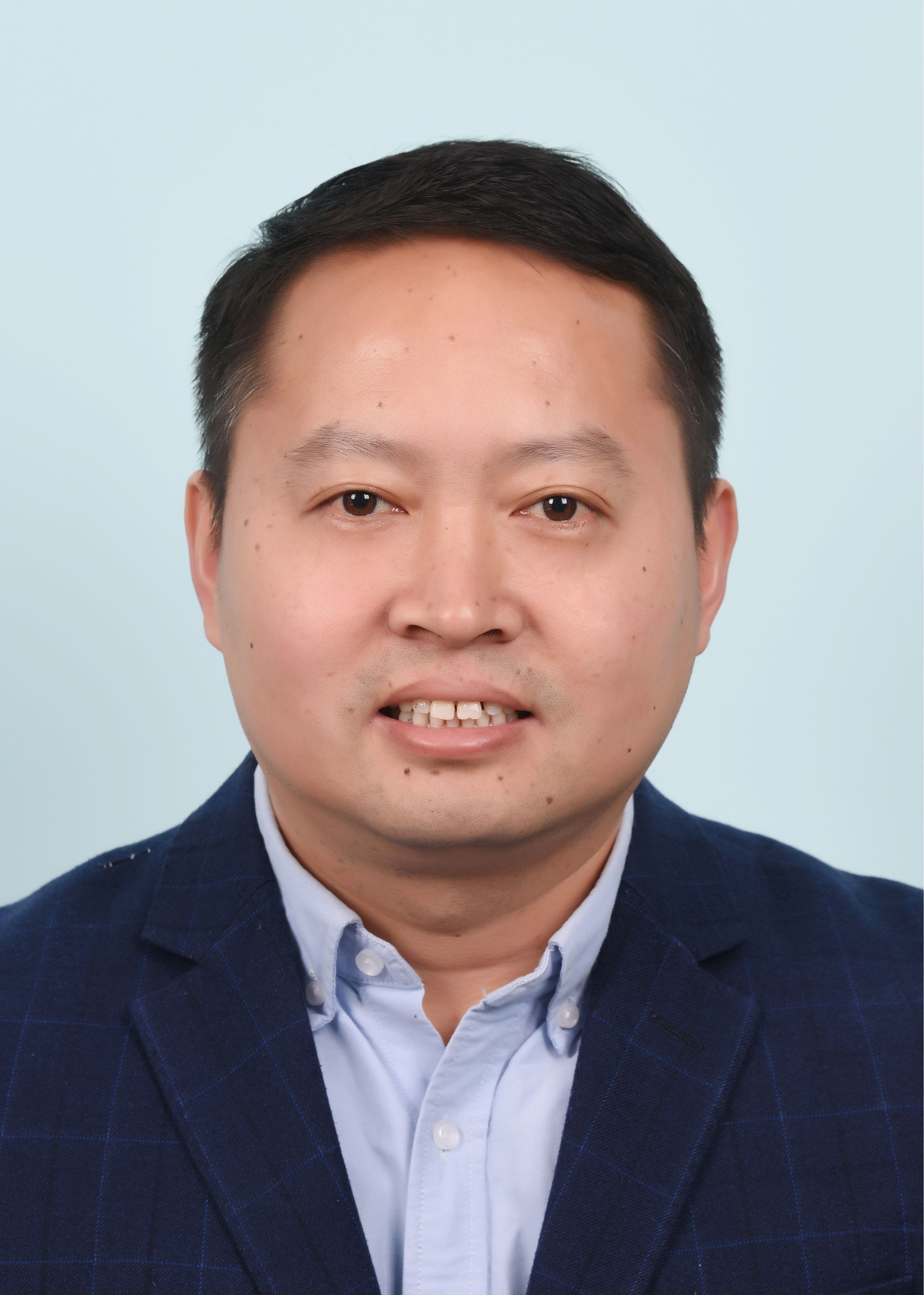}}]{Xianyu Zuo}received the B.S. degree and the M.S. degree in applied mathematics from Henan Normal University, Xinxiang, China, in 2003 and 2006, respectively. He received a Ph.D. degree in computational mathematics from China Academy of Engineering Physical, Xianyang, China, in 2012. He is currently a Professor at the College of Computer Science and Information Engineering of Henan University. His current research interests include high-productivity computing and parallel computing of remote sensing.\end{IEEEbiography}

\end{document}